\documentclass[a4paper,fleqn]{cas-dc}

\usepackage[numbers]{natbib}
\usepackage{algorithm}
\usepackage{algpseudocode}
\usepackage{float}
\usepackage{subcaption}  
\usepackage{xcolor}  
\usepackage{multirow}  
\usepackage{cuted}  

\newcommand{\inc}[1]{\textcolor{green!50!black}{↑ #1}}  
\newcommand{\dec}[1]{\textcolor{red}{↓ #1}}  

\def\tsc#1{\csdef{#1}{\textsc{\lowercase{#1}}\xspace}}
\tsc{WGM}
\tsc{QE}
\tsc{EP}
\tsc{PMS}
\tsc{BEC}
\tsc{DE}

\begin{document}
\let\WriteBookmarks\relax


\setcounter{topnumber}{5}
\setcounter{bottomnumber}{5}
\setcounter{totalnumber}{10}
\setcounter{dbltopnumber}{5}

\renewcommand{\topfraction}{0.9}
\renewcommand{\bottomfraction}{0.8}
\renewcommand{\textfraction}{0.07}      
\renewcommand{\floatpagefraction}{0.5}
\renewcommand{\dbltopfraction}{0.9}
\renewcommand{\dblfloatpagefraction}{0.7}

\setlength{\floatsep}{10pt plus 2pt minus 2pt}
\setlength{\textfloatsep}{10pt plus 2pt minus 2pt}
\setlength{\intextsep}{10pt plus 2pt minus 2pt}
%


\shorttitle{Multi-Agent Reinforcement Learning for Text-to-Image Generation}
\shortauthors{Shi et~al.}

\title [mode = title]{Collaborative Text-to-Image Generation via Multi-Agent Reinforcement Learning and Semantic Fusion}                      


\author[1]{Jiabao Shi}
\affiliation[1]{organization={Minzu University of China},
                city={Beijing},
                country={China}}

\author[2]{Minfeng Qi}
\cormark[1]
\affiliation[2]{organization={City University of Macau},
                city={Macau SAR},
                country={China}}

\author[2]{Lefeng Zhang}

\author[3,2]{Di Wang}
\affiliation[3]{organization={Key Laboratory of Computing Power Network and Information Security, Ministry of Education, Shandong Computer Science Center (National Supercomputer Center in Jinan), Qilu University of Technology (Shandong Academy of Sciences)},
                city={Jinan},
                country={China}}

\author[1]{Yingjie Zhao}
\author[2]{Ziying Li}

\author[2]{Yalong Xing}

\author[4]{Ningran Li}

\affiliation[4]{organization={The University of Adelaide},
                city={Adelaide},
                country={Australia}}

\cortext[cor1]{Corresponding author: Minfeng Qi (mfqi@cityu.edu.mo)}

\begin{abstract}
Multimodal text-to-image generation remains constrained by the difficulty of maintaining semantic alignment and professional-level detail across diverse visual domains. We propose a multi-agent reinforcement learning framework that coordinates domain-specialized agents (e.g., targeting architecture, portraiture, and landscape imagery) within two coupled subsystems: a text enhancement module and an image generation module, each augmented with multimodal integration components. Agents are trained using Proximal Policy Optimization (PPO) under a composite reward function that balances semantic similarity, linguistic/visual quality, and content diversity. Cross-modal alignment is enforced through contrastive learning, bidirectional attention, and iterative feedback between text and image. Across six experimental settings, our system significantly enriches generated content (word count +1614\%) while reducing ROUGE-1 scores by 69.7\%. Among fusion methods, Transformer-based strategies achieve the highest composite score (0.521) despite occasional stability issues. Multimodal ensembles yield moderate consistency (0.444–0.481), reflecting the persistent challenges of cross-modal semantic grounding. These findings underscore the promise of collaborative, specialization-driven architectures for advancing reliable multimodal generative systems.
\end{abstract}




\maketitle

\section{Introduction}

The rapid progress of multimodal generation has produced models such as GPT-4 and DALL-E that exhibit impressive zero-shot capabilities across diverse domains. Yet, these systems reveal a fundamental tension between broad generalization and domain-specific precision. While they can generate content across a wide variety of contexts, their outputs often fail to meet the nuanced professional standards demanded in specialized areas such as architectural visualization, portraiture, or landscape design. This gap reflects a structural limitation of monolithic architectures: one model attempting to master all domains inevitably sacrifices depth of expertise. The challenge is not merely achieving acceptable quality across domains, but rather maintaining professional-grade precision in each specialized area—a requirement that becomes increasingly difficult as model capacity is distributed across heterogeneous tasks.

A closer look at professional creative practice underscores this limitation. Human content creators rarely rely on a single tool for all tasks; instead, they employ domain-specific workflows tailored to specialized requirements. Architectural visualization demands structural accuracy and technical terminology, whereas portrait generation emphasizes facial anatomy and compositional aesthetics. Landscape design, in turn, requires careful attention to environmental elements, lighting conditions, and atmospheric effects. These contrasting needs suggest that a collaborative, multi-expert approach may better mirror human creative processes than current single-model solutions. The division of labor inherent in professional workflows points toward a fundamental organizational principle: specialization enables depth of expertise that generalist approaches cannot easily or efficiently replicate.

At the same time, advances in multimodal fusion have primarily targeted "understanding" tasks rather than "generation". Existing text-to-image systems that attempt multi-agent designs often lack principled coordination mechanisms and do not leverage reinforcement learning for iterative refinement. As a result, they struggle to maintain cross-modal semantic alignment, ensuring that generated images remain faithful to textual descriptions throughout the generation process. This disconnect between textual intent and visual output represents a critical bottleneck in current multimodal generation pipelines, particularly when professional applications demand strict adherence to descriptive specifications and requirements.

\smallskip
\noindent\textbf{Research gaps.}  
We identify three critical shortcomings in the current landscape. First, single-agent architectures exhibit performance degradation when applied to heterogeneous domains, lacking the professional accuracy expected in specialized applications. When a monolithic model must balance architectural precision, portrait fidelity, and landscape aesthetics within limited capacity, it typically achieves mediocre performance across all domains rather than excellence in any. Second, existing multi-agent approaches remain ad hoc, with no systematic mechanisms for coordination or reinforcement-driven quality improvement. The absence of learnable coordination strategies means that agent outputs are often combined through simple heuristics, failing to capture the nuanced interactions between domain-specific expertise. Third, evaluation frameworks inadequately capture multimodal performance, relying on single-modality metrics and neglecting semantic alignment across text and images. This fragmented assessment obscures the true challenge of multimodal generation: maintaining semantic coherence and quality simultaneously across both modalities. Collectively, these limitations impede both technical progress and practical deployment in professional contexts such as design, education, and marketing.

\smallskip
\noindent\textbf{Our Design.}  
To address these shortcomings, we develop a multi-agent reinforcement learning framework for multimodal generation. The system organizes domain-specific agents specialized in architecture, portraiture, and landscape tasks, coordinated under a unified PPO-based optimization scheme. Each agent is equipped with domain-tailored knowledge and generation capabilities, allowing the system to preserve professional precision while maintaining flexibility across diverse content types. Cross-modal alignment is enforced through specialized fusion modules, while a comprehensive evaluation framework integrates linguistic, visual, and semantic measures. This design explicitly balances specialization with generalization and emphasizes semantic fidelity between modalities. By decomposing the generation pipeline into coordinated expert modules, the framework addresses the core limitations of monolithic systems while introducing learnable mechanisms for quality optimization and cross-modal consistency.

\smallskip
\noindent\textbf{Contributions.}  
This paper makes the following contributions:

\noindent\hangindent 1em\textit{$\triangleright$ A specialized multi-agent architecture for multimodal generation.}  
We propose a domain-aware design where expert agents (e.g., architecture, portrait, landscape) collaborate through coordination mechanisms, surpassing monolithic single-agent systems in content precision and professional rigor. This architectural choice directly addresses the specialization-generalization trade-off by distributing domain expertise across purpose-built agents.

\noindent\hangindent 1em\textit{$\triangleright$ Reinforcement learning for cross-modal optimization.}  
We integrate Proximal Policy Optimization (PPO) into the training process to explore joint optimization of text and image quality metrics. Our empirical analysis reveals modality-specific learning dynamics, with PPO showing effectiveness in image generation while encountering challenges in text generation due to multi-agent non-stationarity.

\noindent\hangindent 1em\textit{$\triangleright$ Advanced multimodal fusion strategies.}  
We implement and evaluate multiple fusion schemes (e.g., dynamic weighting, Transformer-based fusion, neural fusion) that strengthen semantic alignment while addressing efficiency–effectiveness trade-offs. Through systematic comparison, we characterize the practical advantages and limitations of each approach in maintaining cross-modal consistency.

\noindent\hangindent 1em\textit{$\triangleright$ A comprehensive evaluation framework.}  
We develop a unified assessment protocol covering text quality, image fidelity, and cross-modal consistency, and validate our framework through extensive experiments that reveal both capabilities and practical limitations of the proposed system. This holistic evaluation addresses the fragmentation of existing metrics and provides a foundation for assessing multimodal generation quality.

\section{Related work}

\subsection{Multi-agent reinforcement learning and PPO optimization}

In recent years, multi-agent reinforcement learning has shown significant advantages in dealing with complex environments where collaboration and competition coexist, PPO has become the mainstream choice due to its training stability and simplicity of implementation. Yu et al.~\cite{balanced_ppo_2025} proposed a balanced PPO method to address the problem of uneven capabilities between weak and strong agents in multi-agent systems, achieving a fairer learning process through gradient reweighting and loss function adjustment. The successful practice of multi-threaded PPO in robot grasping tasks has demonstrated the stability of the algorithm under limited computing power and real-time constraints~\cite{robot_arm_ppo_2024}, while the UAV spectrum scheduling scheme based on PPO-assisted evolutionary reinforcement learning has verified its robustness and transferability in incomplete information environments~\cite{uav_spectrum_2024}. Ma et al.~\cite{multi_scale_ppo_2025} pointed out the effectiveness of PPO in multi-objective optimization scenarios in gradient imbalance analysis.

However, the existing PPO method lacks targeted design in text-to-image multimodal content generation, especially in terms of semantic consistency and cross-domain professional quality output, which faces major gaps and challenges. Based on these limitations, this paper constructs two interrelated multi-agent subsystems consisting of a text augmentation system and an image generation system. Each system contains domain-specific agents specialized for architecture, portraiture, and landscape, and is equipped with a multimodal integration module, aiming to address the limitations of single-agent text-to-image generation systems.

\subsection{Cooperative Control and Consensus Theory }

The cooperative control theory of multi-agent systems provides an important theoretical basis for the dual-subsystem architecture proposed in this paper. For the controllability problem of discrete-time multi-agent systems under matrix-weighted networks, existing studies have established controllability criteria based on the Graph Laplast property~\cite{controllability_2023}. The study of interval consistency of heterogeneous high-order multi-agent systems has further expanded the scope of application of cooperative control, and the Lyapunov-Krasovskii method is used to deal with time-varying parameters and communication delay problems~\cite{interval_consensus_2025}. In terms of robustness, H2 robust consistency analysis under parameter uncertainty and finite time lag consistency control based on passivity provide verifiable convergence guarantees for dealing with model errors and external disturbances~\cite{passivity_consensus_2024} ~\cite{robust_h2_2024}.

However, these theoretical studies mainly focus on the optimization of a single control objective and lack a systematic analysis of multi-objective cooperative optimization (such as comprehensive evaluation of similarity, quality and diversity). In particular, the strict feedback nonlinear multi-agent reinforcement learning method based on backstepping design~\cite{game_backstepping_2024} provides a structural design idea for complex feedback loops, but there is still a theoretical gap in cross-modal semantic alignment. To address these shortcomings, this paper proposes a multi-agent reinforcement learning framework through specialized agent collaboration and advanced fusion mechanism, PPO is used for agent training, a comprehensive reward function is used to evaluate similarity, quality and diversity indicators, and cross-modal alignment is achieved through consistency evaluation, iterative parameter optimization and bidirectional text-image feedback loop.

\subsection{Multimodal fusion and Transformer architecture}

The development of multimodal fusion technology has provided strong technical support for cross-domain information integration~\cite{medical_fusion_2012}, which is directly related to the design of the multimodal integration module in this paper. The dual-branch CNN-Transformer feature extraction network (DBCTfuse) proposed by Li et al.~\cite{dbctfuse_2024} achieved excellent fusion effect by processing different modal features in parallel, demonstrating the advantages~\cite{rgb_depth_fusion_2024} of deep learning methods in infrared and visible light image fusion. The Image Fusion Transformer proposed by Vibashan et al.~\cite{image_fusion_transformer_2022} marked a breakthrough in the pure Transformer architecture in image fusion tasks, and the MATCNN method based on multi-scale CNN and attention Transformer by Tang et al.~\cite{matcnn_2025} further demonstrated the effectiveness of the hybrid architecture. The TCUP-Fusion method for fusion of ultrasound and photoacoustic images~\cite{tcup_fusion_2024} demonstrated the potential of combining Transformer and CNN in medical multimodal fusion  applications.

Although these studies have made significant progress in fusion technology, they still face challenges in numerical stability and API compatibility in practical applications, especially the lack of targeted fusion mechanism design in complex multimodal generation tasks. Given these technical challenges, our experiments demonstrate the potential and practical challenges of multi-agent approaches. Among the tested fusion methods, the Transformer achieves the highest performance despite numerical stability challenges.

\subsection{Text-to-image generation and multimodal content creation}

The latest progress in text-to-image generation technology provides a direct technical comparison benchmark for the core task of this paper. Harzig et al.~\cite{text_image_parallel_2024} pioneered an end-to-end multimodal generation paradigm based on the text-to-image generation method of parallel decoding of vision-language Transformer. Its parallel processing mechanism echoes the dual-subsystem design concept of this paper. Optimization of text-to-image generation for accurate training-independent glyph enhancement~\cite{text_image_glyph_2025} pays special attention to the text accuracy of generated content, which is highly relevant to the method of solving the semantic consistency challenge through a bidirectional text-image feedback loop in this paper. Faidon et al.~\cite{multimodal_text_classification_2023} demonstrated the advantages of Transformer in handling text-image association tasks through multimodal multi-label text classification based on bidirectional Transformer~\cite{agents_image_extraction_2021}.

However, existing research still faces significant challenges in multimodal semantic alignment, especially in achieving cross-modal consistency and avoiding object hallucination~\cite{mousse_2025}. To address these shortcomings, this paper provides a comprehensive technical implementation framework for identifying key challenges in future multi-agent generation systems and contributing valuable insights into the development of collaborative AI in the creative field.

\section{Methodology}

\subsection{System Architecture Overview}

Our framework adopts a modular decomposition strategy to address the fundamental limitations of monolithic text-to-image generation systems. A single large model that expands prompts tends to dilute domain-specific terminology and overlook structural rigor in professional settings. To counter this, we decompose the pipeline into three domain-grounded modules—text enhancement, image generation, and multimodal integration—each optimized for a distinct stage of the generation process.

\begin{figure*}[t]
    \centering
    \includegraphics[width=0.9\textwidth]{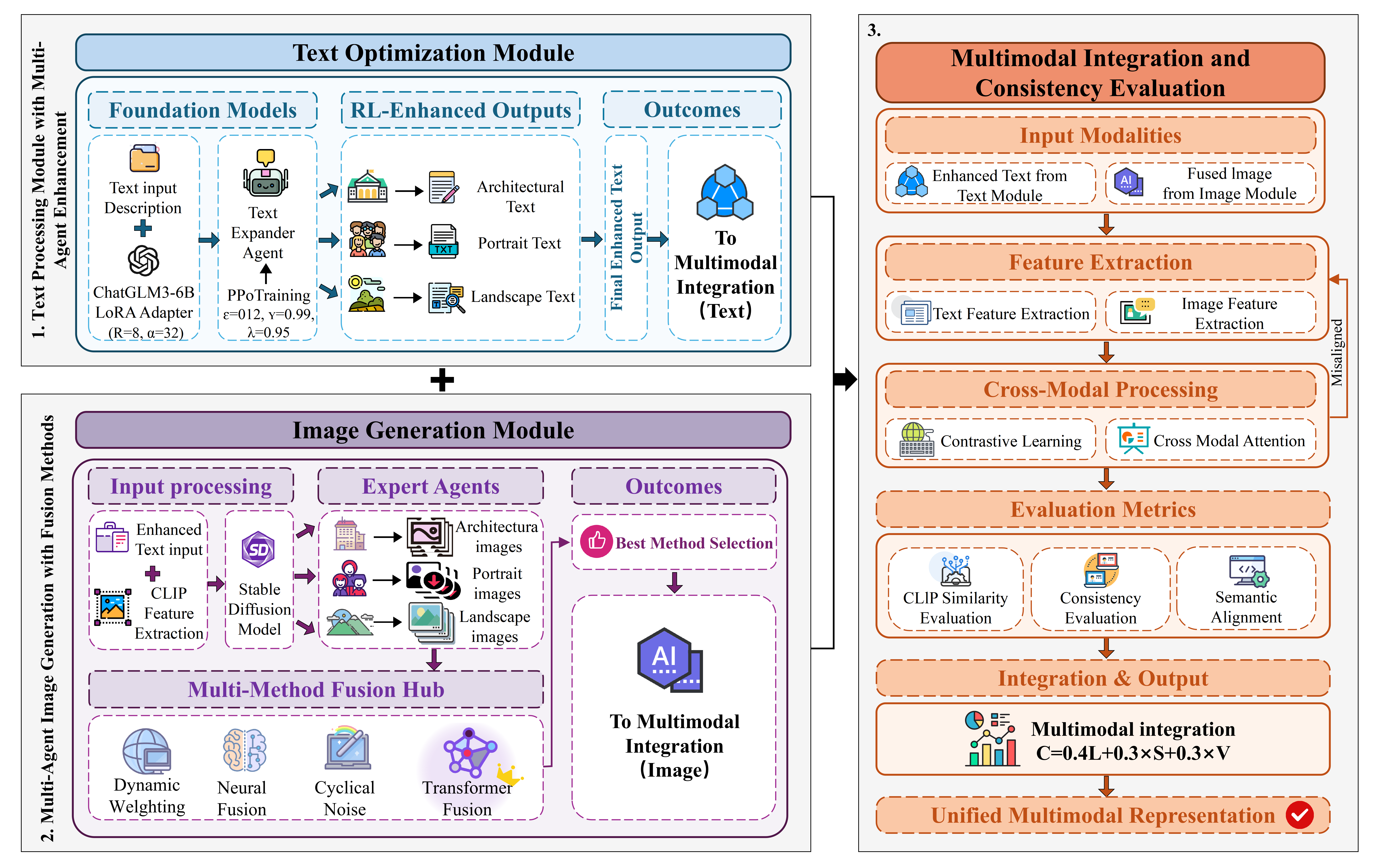}
    \caption{Multi-agent Text-to-Image Generation Model framework diagram. The system comprises three main modules: Text Optimization Module (left) with foundation models and enhancement agents, Image Generation Module (bottom) with multi-method fusion strategies, and Multimodal Integration and Consistency Evaluation Module (right).}
    \label{fig:flowchart}
\end{figure*}

This modular design addresses key limitations by isolating concerns, preserving professional vocabulary and structure, and enabling targeted optimization without destabilizing the entire system. A lightweight coordination mechanism maintains coherence across modules and ensures information flows consistently from text to image and then to cross-modal checking~\cite{tang2020wheel}. The architecture employs specialized agents at each stage: text agents enhance prompts with domain expertise, image agents generate visual content with professional accuracy, and integration modules enforce semantic consistency across modalities. The complete end-to-end pipeline is depicted in Figure~\ref{fig:flowchart}, showing how these components interact to produce semantically consistent multimodal outputs.
\subsection{Text Enhancement Framework}

The text enhancement module addresses a critical challenge: single-model prompt expansion tends to produce generic descriptions that lose professional specificity and domain expertise. To counteract this, we deploy four specialized agents—expander, architecture, portrait, and landscape—that work collaboratively to enrich user prompts while preserving intent and injecting domain precision where it matters most (terminology, structural constraints, and visual attributes)~\cite{counterfactual_text_2024}.
\begin{center}
    \includegraphics[width=0.4\textwidth]{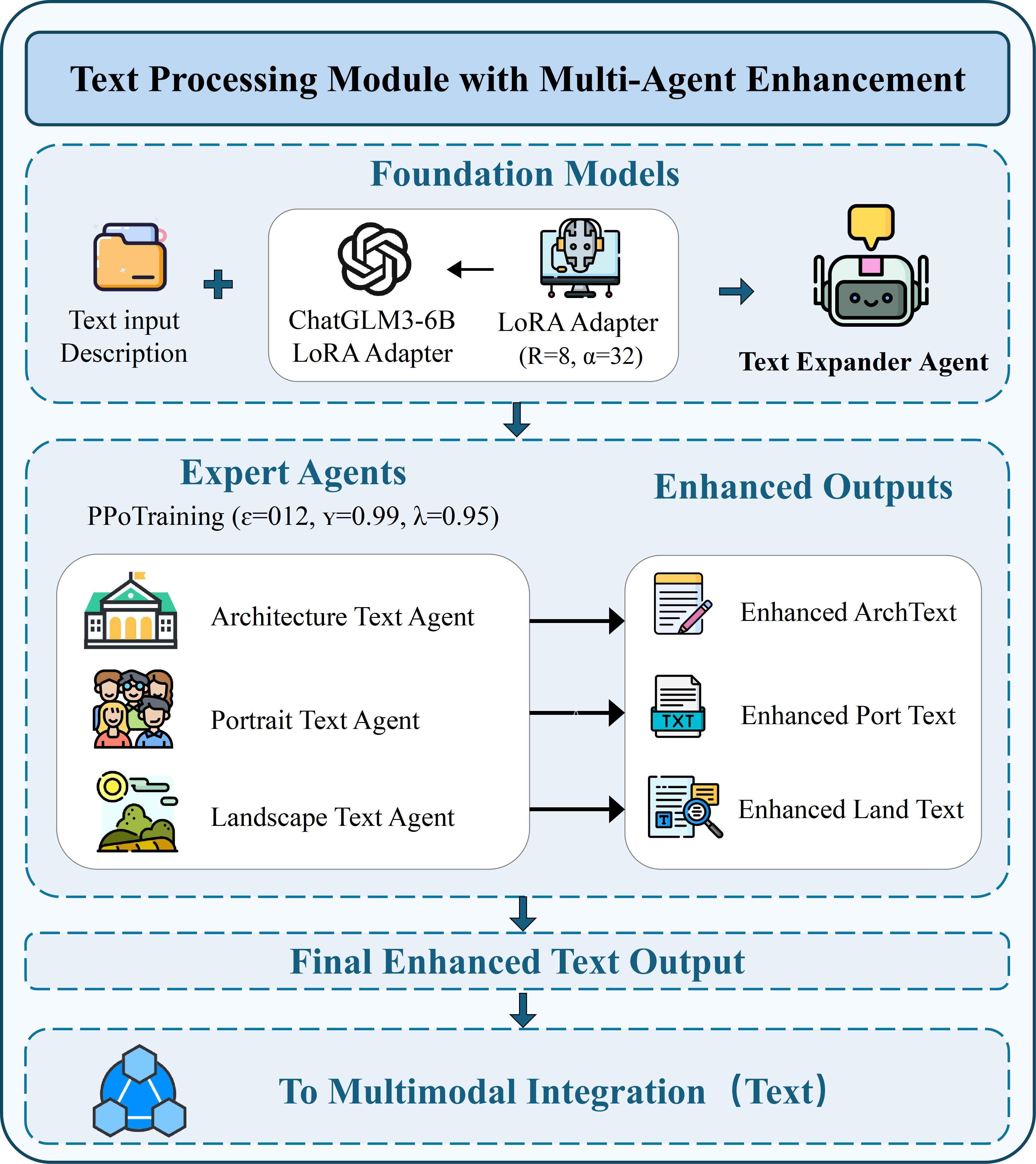}
    \captionof{figure}{Text Processing Module with Multi-Agent Enhancement}
    \label{fig:text_process}
\end{center}
Each agent is built upon a pre-trained language model enhanced with parameter-efficient fine-tuning (PEFT) using LoRA adapters configured as: rank = 8, $\alpha = 32$, dropout = 0.1~\cite{lstm_text_generation_2020}. This configuration balances adaptation capacity with training efficiency. The expander agent performs general-purpose elaboration, while the three domain-specific agents contribute specialized knowledge for their respective domains. The architectural workflow of this multi-agent enhancement process is visualized in Figure~\ref{fig:text_process}.

We adopt proximal policy optimization (PPO) to jointly improve fluency and specificity. The clipped objective is
\begin{equation}
L^{\text{CLIP}}(\theta) = \mathbb{E}_t\!\left[\min\!\big(r_t(\theta)\hat{A}_t,\ \text{clip}(r_t(\theta), 1-\epsilon, 1+\epsilon)\hat{A}_t\big)\right],
\end{equation}
where $r_t(\theta)=\frac{\pi_\theta(a_t|s_t)}{\pi_{\theta_{\text{old}}}(a_t|s_t)}$. The reward combines multiple indicators:
\vspace{-4pt}
\begin{align}
R(D, R_{ref}) &= 0.4 \cdot \text{BLEU} + 0.3 \cdot \text{ROUGE} \nonumber \\
&\quad + 0.2 \cdot \text{Coherence} + 0.1 \cdot \text{Diversity}.
\end{align}
In practice, PPO is run with standard settings (e.g., $\epsilon=0.2$, $\gamma=0.99$), stabilizing updates while preventing policy collapse. This module outputs enriched, domain-faithful prompts that feed directly into the image generation stage for style- and scene-specific synthesis.

The detailed steps of the PPO training algorithm are as follows in Algorithm~\ref{alg:ppo_training}:


\subsection{Image Generation Framework}

The image generation module employs three specialized visual agents—architecture, portrait, and landscape—to preserve professional accuracy across diverse visual domains. This specialization is motivated by a fundamental observation: monolithic generators often generalize away crucial domain-specific details such as structural elements in architectural scenes, facial fidelity in portrait images~\cite{fukuda2011facial}, and environmental lighting and weather conditions in landscape views~\cite{liu2024weather}. By distributing generation responsibilities across domain experts, we maintain the precision required for professional applications.

We build upon Stable Diffusion as the base generative engine~\cite{tosa2025chromagazer}, enhanced with agent-specific latent conditioning derived from the enriched text features. The generation pipeline operates through four sequential stages: (1) text feature extraction using CLIP encoders, (2) routing features through agent-specific layers to obtain specialized latent representations, (3) parallel candidate image generation by each agent based on its latent conditioning, and (4) multi-agent fusion to produce the final output.

We support four fusion strategies to accommodate different quality and efficiency requirements: weighted average, attention-based weighting, transformer fusion, and simple averaging. The fused result, along with intermediate features, is then forwarded to the multimodal integration module for comprehensive cross-modal alignment and detailed consistency checking. The complete architecture and dataflow are shown in Figure~\ref{fig:image_process}.

\begin{center}
    \includegraphics[width=0.4\textwidth]{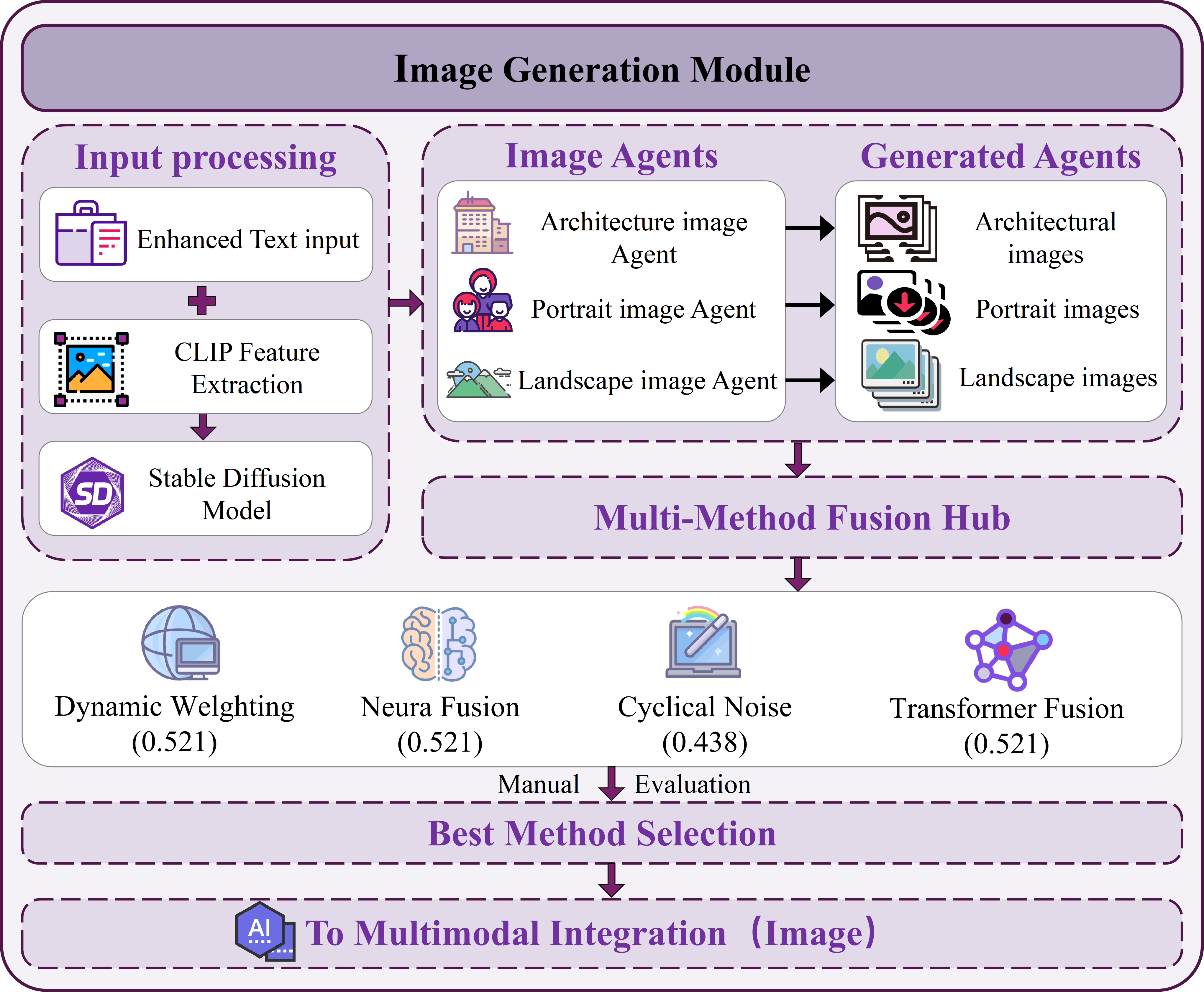}
    \captionof{figure}{Multi-Agent Image Generation with Fusion Methods}
    \label{fig:image_process}
\end{center}

\subsection{Advanced Fusion Methods}

We implement multiple fusion strategies to accommodate different content structures and quality constraints. The design space includes: transformer-based fusion~\cite{nie2024cross} (multi-head attention to learn content-aware combination weights), dynamic weighted fusion (weights adapt to estimated quality and content complexity), neural fusion~\cite{xia2024intelligent} (learning non-linear mappings from data), content-aware fusion (semantic cues guide composition), and simple averaging (a robust baseline)~\cite{zhong2024unsupervised}. The aim is not to enumerate algorithmic minutiae, but to provide a spectrum from lightweight baselines to learned, content-adaptive schemes so the system can select an appropriate trade-off between fidelity, smoothness, and efficiency. The fused image is then evaluated and refined under the multimodal integration module to ensure text–image agreement before reporting final outputs.






\subsection{Multimodal Integration}

\begin{table*}[t]
\centering
\caption{Comprehensive Multi-Agent System Performance Evaluation.}
\label{tab:multiagent_performance}
\scriptsize
\renewcommand{\arraystretch}{0.92}
\setlength{\tabcolsep}{3pt}
\resizebox{\textwidth}{!}{
\begin{tabular}{l|l>{\columncolor{gray!10}}c>{\columncolor{gray!10}}c>{\columncolor{gray!10}}c>{\columncolor{gray!10}}c>{\columncolor{gray!10}}c l}
\toprule
\textbf{Experiment} & \textbf{Scenario/Method} & \textbf{Metric 1} & \textbf{Metric 2} & \textbf{Metric 3} & \textbf{Metric 4} & \textbf{Metric 5} & \textbf{Improvement} \\
\midrule
\multirow{5}{*}{Text Generation} 
    & Medieval castle       & 1.000 & 1.000 & 0.769 & 0.286 & 4.7s   & \inc{0.0\%}     \\
    & Modern city           & 1.000 & 0.833 & 0.800 & 0.667 & 4.4s   & \dec{16.7\%}   \\
    & Elderly wise man      & 1.000 & 0.429 & 0.824 & 0.071 & 4.4s   & \dec{57.1\%}   \\
    & Mountain landscape    & 1.000 & 0.500 & 0.727 & 0.042 & 4.6s   & \dec{50.0\%}   \\
    & Space station         & 1.000 & 1.000 & 0.727 & 0.100 & 4.5s   & \inc{0.0\%}     \\
\midrule
\multirow{3}{*}{PPO Training}
    & Gothic cathedral      & 1.000 & 1.000 & 0.800 & 0.600 & 0.488  & \dec{39.9\%}   \\
    & Enchanted forest      & 1.000 & 1.000 & 0.824 & 0.609 & 0.433  & \dec{45.7\%}   \\
    & Mountain lake         & 1.000 & 1.000 & 0.800 & 0.571 & 0.456  & \dec{43.1\%}   \\
\midrule
\multirow{3}{*}{Image Generation}
    & Mountain castle       & 0.446 & 0.456 & 0.700 & 0.700 & N/A    & \inc{2.2\%}    \\
    & Noble knight          & 0.446 & 0.456 & 0.650 & 0.650 & N/A    & \inc{2.2\%}    \\
    & Forest clearing       & 0.446 & 0.456 & 0.650 & 0.650 & N/A    & \inc{2.2\%}    \\
\midrule
\multirow{2}{*}{Multimodal Opt.}
    & Medieval fortress     & 0.650 & 0.820 & 0.710 & 0.880 & Enhanced & \inc{26.2\%} \\
    & Crystal palace        & 0.650 & 0.820 & 0.710 & 0.880 & Enhanced & \inc{26.2\%} \\
\midrule
\multirow{4}{*}{Fusion Methods}
    & Transformer           & 0.417 & 0.521 & 0.625 & 0.625 & 0.003s & Best Quality  \\
    & Neural Fusion         & 0.417 & 0.521 & 0.625 & 0.625 & 0.002s & Best Quality  \\
    & Dynamic Weight        & 0.417 & 0.521 & 0.625 & 0.625 & 0.009s & Best Quality  \\
    & Content Aware         & 0.250 & 0.438 & 0.625 & 0.625 & 0.014s & Moderate Quality  \\
\midrule
\multirow{3}{*}{Integration}
    & Text$\rightarrow$Image - Gothic   & 0.460 & 0.777 & 0.000 & 0.700 & Consistent & Good Alignment  \\
    & Image$\rightarrow$Text - Elderly  & 0.445 & 0.727 & 0.000 & 0.700 & Consistent & Good Alignment \\
    & Bidirectional - Gothic            & 0.445 & 0.729 & 0.000 & 0.700 & Consistent & Good Alignment \\
\bottomrule
\end{tabular}}
\end{table*}

The multimodal integration and consistency evaluation module serves as the critical bridge ensuring deep semantic alignment between textual descriptions and visual representations. This module operates through three complementary and synergistic mechanisms that work in concert to validate and refine cross-modal correspondence. The comprehensive architecture and workflow of this integration process are illustrated in Figure~\ref{fig:multimodal}. 

\begin{center}
    \includegraphics[width=0.4\textwidth]{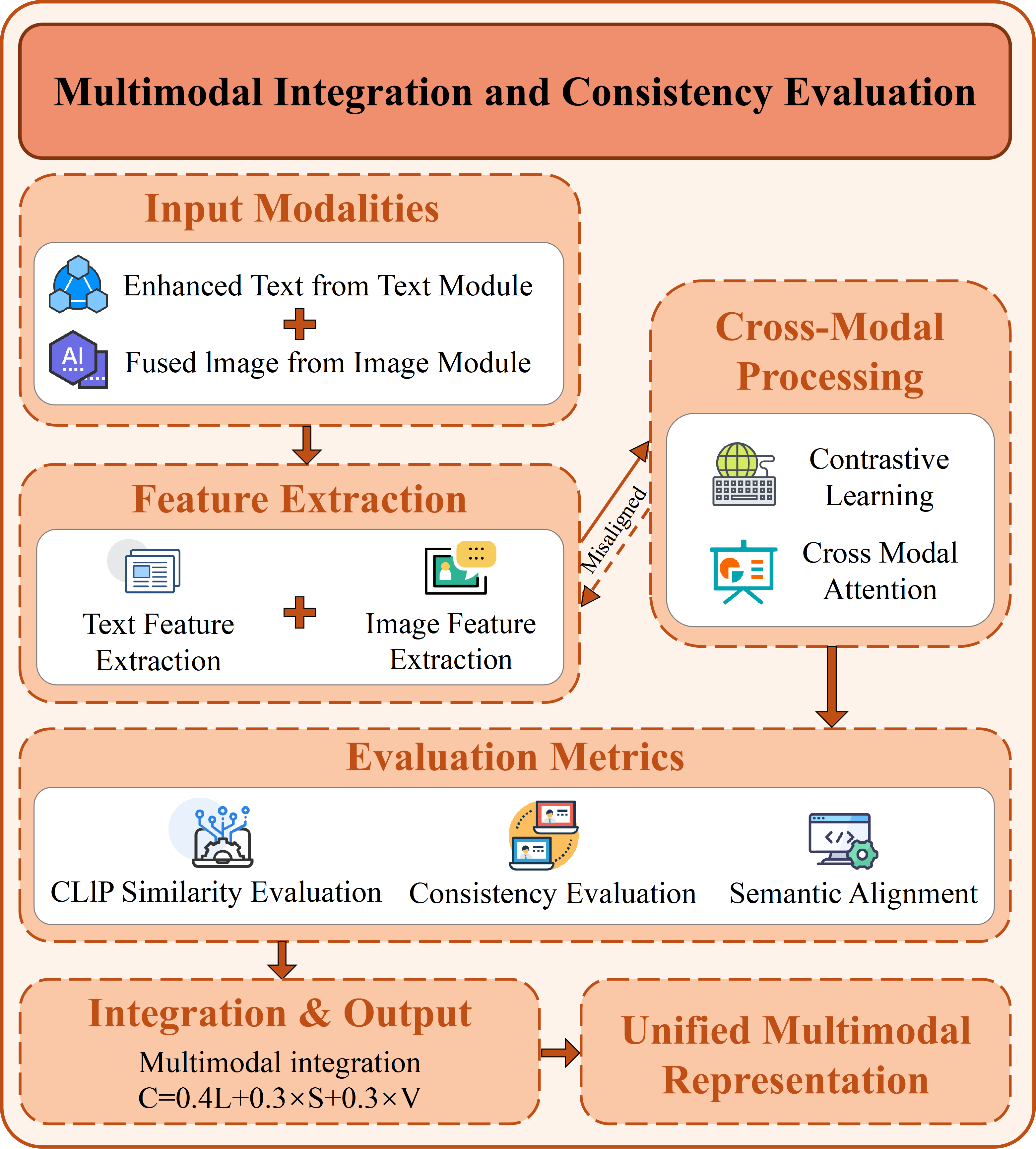}
    \captionof{figure}{Multimodal Integration and Consistency Evaluation}
    \label{fig:multimodal}
\end{center}

The first mechanism employs contrastive alignment in a shared embedding space, utilizing a temperature-scaled objective function that encourages matching text-image pairs to exhibit higher similarity scores than non-matching pairs:
\begin{equation}
\mathcal{L}_{\text{contrastive}} = -\log\frac{\exp(\text{sim}(z_t, z_i)/\tau)}{\sum_{j}\exp(\text{sim}(z_t, z_j)/\tau)},
\end{equation}
where $z_t$ and $z_i$ denote the embedded representations of text and image respectively, $\tau$ controls the temperature parameter for scaling the similarity distribution, and the summation iterates over all candidate image embeddings in the batch.

The second mechanism implements bidirectional cross-modal attention~\cite{liu2025semantic}, which captures fine-grained dependencies and correspondences between individual textual elements (such as words, phrases, and semantic units) and specific visual regions (including objects, textures, and spatial locations) within the generated images. This attention mechanism operates in both directions: text-to-image attention identifies which visual regions correspond to each textual element, while image-to-text attention determines which textual descriptions are most relevant and accurate to each visual feature.

The third mechanism computes a composite consistency score that aggregates multiple complementary alignment signals into a single unified metric. This composite score integrates three distinct components with carefully calibrated weights: contrastive alignment loss (weighted at 0.4), cosine similarity between embeddings (weighted at 0.3), and object/keyword validation accuracy (weighted at 0.3):
\begin{equation}
C \;=\; 0.4 \cdot \big(1 - \mathcal{L}_{\text{contrast}}\big)\;+\;0.3 \cdot \text{cos\_sim}\;+\;0.3 \cdot \text{obj\_valid}.
\end{equation}
These weights reflect the relative importance and reliability of each component in assessing overall multimodal consistency and alignment quality.

The integration module produces two primary outputs: a semantically refined image that incorporates multimodal consistency constraints, and a set of alignment scores that quantify cross-modal correspondence quality. These alignment scores serve as interpretable indicators of text-image consistency and inform downstream quality assessment. The complete computational procedure for evaluating multimodal consistency is formalized in Algorithm~\ref{alg:multimodal_consistency}, which details the sequential steps for feature normalization, similarity computation, and composite score aggregation:

\begin{algorithm}[!t]
\caption{Multimodal Consistency Assessment}
\label{alg:multimodal_consistency}
\begin{algorithmic}[1]
\Require Text features $F_t$, Image features $F_i$, Temperature parameter $\tau$
\Ensure Consistency score $C$

\State \textbf{Normalize features:} $\hat{F}_t \gets \frac{F_t}{\|F_t\|_2}$, $\hat{F}_i \gets \frac{F_i}{\|F_i\|_2}$
\State \textbf{Compute similarity matrix:} $S_{ij} \gets \frac{\hat{F}_{t,i} \cdot \hat{F}_{i,j}}{\tau}$
\State \textbf{Compute contrastive loss:} $\mathcal{L}_{\text{contrast}} \gets -\log\left(\frac{\exp(S_{ii})}{\sum_j \exp(S_{ij})}\right)$
\State \textbf{Compute cosine similarity:} $\text{cos\_sim} \gets \frac{\hat{F}_t \cdot \hat{F}_i}{\|\hat{F}_t\| \cdot \|\hat{F}_i\|}$
\State \textbf{Keyword detection:} $K_t \gets \text{ExtractKeywords}(T)$,\quad $K_i \gets \text{ObjectDetection}(I)$
\State \textbf{Object validation score:} $\text{obj\_valid} \gets \frac{|K_t \cap K_i|}{|K_t \cup K_i|}$
\State \textbf{Final consistency score:} $C \gets 0.4 \cdot (1 - \mathcal{L}_{\text{contrast}}) + 0.3 \cdot \text{cos\_sim} + 0.3 \cdot \text{obj\_valid}$
\State \Return $C$
\end{algorithmic}
\end{algorithm}

\section{Experimental design and results}

\subsection{Experimental environment}

\smallskip
\noindent\textbf{Software and hardware environment.}
We conduct experiments on Windows 11 with Python 3.8+ and PyTorch (CUDA 11.3 for GPU acceleration). To ensure reproducibility and robustness, we adopt a proven stack: torch ($\ge$2.0.0), transformers ($\ge$4.35.0), accelerate ($\ge$0.20.0), and diffusers ($\ge$0.21.0), together with general-purpose tooling including numpy, pandas, matplotlib, seaborn, scikit-learn, Pillow, and tqdm. For text evaluation we use nltk, rouge-score, sacrebleu, and sentencepiece; bitsandbytes is optionally used for acceleration/quantization. The hardware platform includes an NVIDIA RTX 4060 GPU, a 13th Gen Intel Core i9-13900HX (2.20 GHz) CPU, 32GB RAM, and $\ge$100GB SSD.

\smallskip
\noindent\textbf{Pretrained models.}
For text generation we use ChatGLM3-6B fine-tuned with LoRA. For image generation and understanding we use Stable Diffusion v1-5 (runwayml/stable-diffusion-v1-5), ResNet50, and CLIP ViT-B/32.

\smallskip
\noindent\textbf{Training and inference configuration.}
Batch sizes are 16 (text) and 8 (image); training epochs are 100 (text) and 50 (image). The initial learning rate is 5e-5 with a linear decay schedule. PPO is configured with clipping coefficient $\epsilon=0.2$ and discount factor $\gamma=0.99$. The reward function combines similarity (0.5), quality (0.3), and diversity (0.2).

\begin{figure*}[!t]
    \centering
    \begin{subfigure}[t]{0.48\textwidth}
        \centering
        \includegraphics[width=\linewidth]{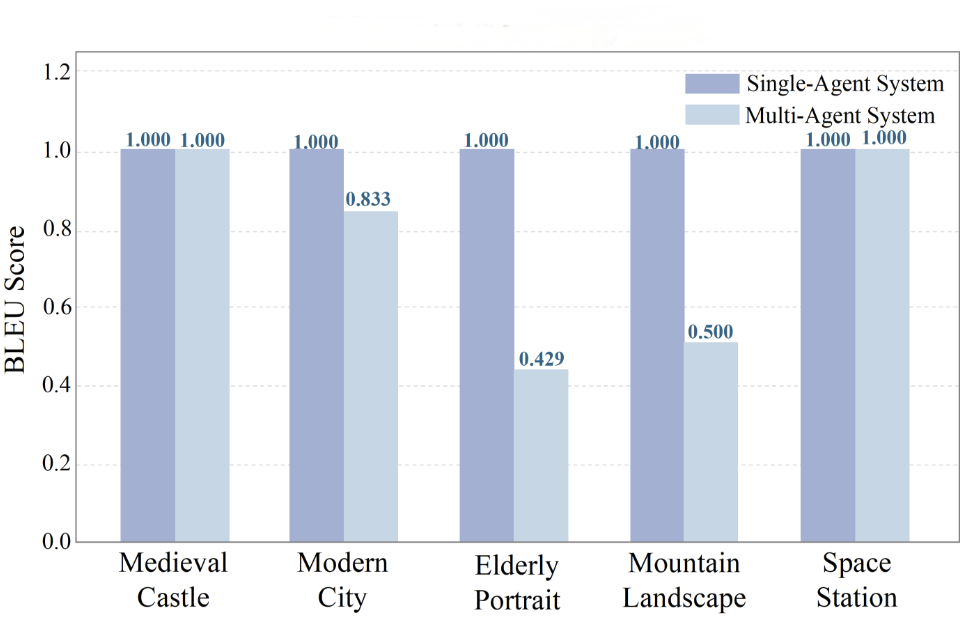}
        \caption{BLUE Score Comparison}
        \label{fig:exp1-1}
    \end{subfigure}
    \hfill
    \begin{subfigure}[t]{0.48\textwidth}
        \centering
        \includegraphics[width=\linewidth]{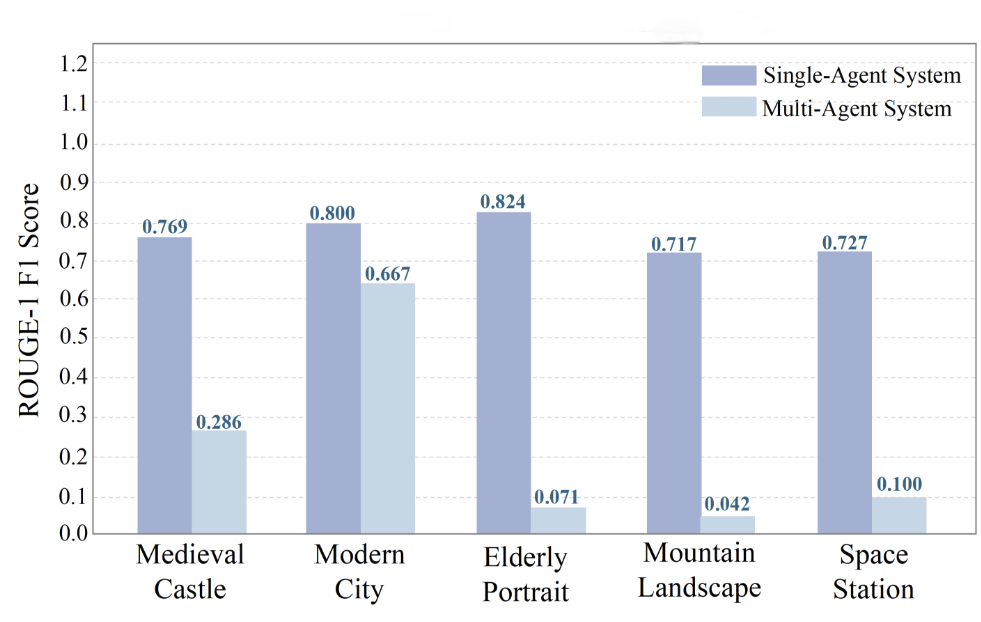}
        \caption{ROUGE-1 F1 Score Comparison}
        \label{fig:exp1-2}
    \end{subfigure}
    
    \vspace{0.5cm} 
    
    \begin{subfigure}[t]{0.48\textwidth}
        \centering
        \includegraphics[width=\linewidth]{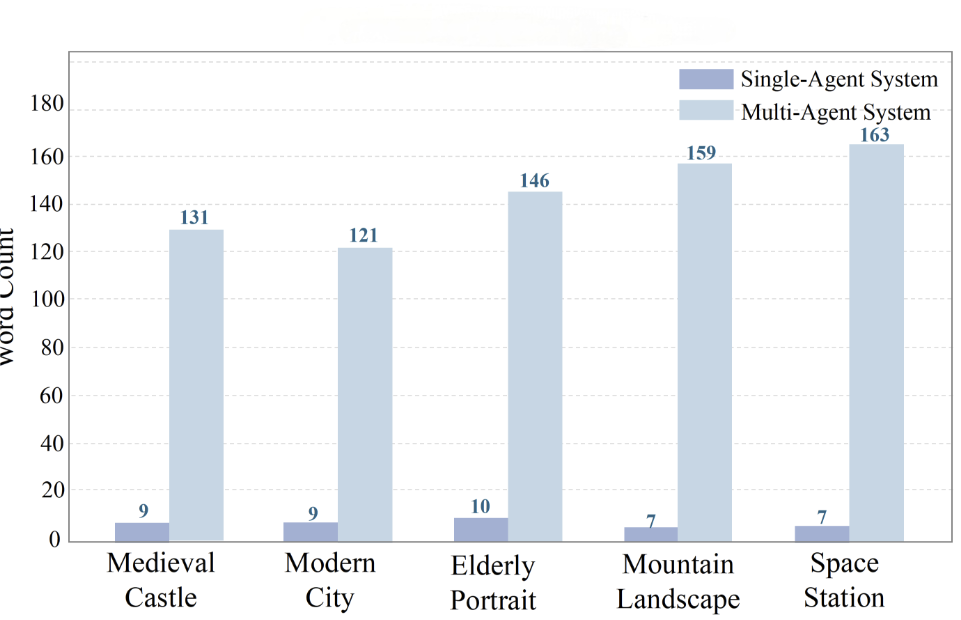}
        \caption{Word Count Comparison}
        \label{fig:exp1-3}
    \end{subfigure}
    \hfill
    \begin{subfigure}[t]{0.48\textwidth}
        \centering
        \includegraphics[width=\linewidth]{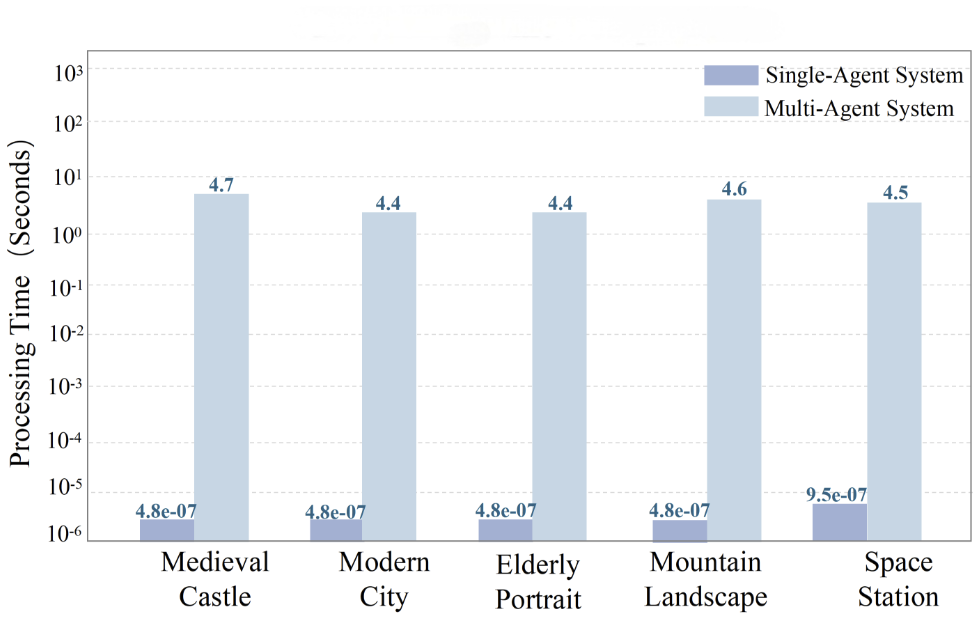}
        \caption{Processing Time Comparison}
        \label{fig:exp1-4}
    \end{subfigure}

    \caption{Single-Agent vs Multi-Agent Text Generation}
    \label{fig:1x4_grid}
\end{figure*}
\subsection{Experimental design}

\noindent\textbf{Rationale for the experiment matrix and ordering.} To systematically assess the capability boundary and practical utility of the proposed multi-agent framework, we organize the experiment matrix along three axes: modality (text/image), learning regime (baseline performance/PPO), and system components (fusion/consistency). The ordering follows a principled progression: we first compare single vs. multi-agent in text (Exp. 1), then examine PPO effects in the same domain (Exp. 2); we transition to image generation with single vs. multi-agent (Exp. 3), followed by PPO effects in image (Exp. 4); we then evaluate core cross-modal fusion techniques (Exp. 5), and finally assess multimodal consistency (Exp. 6). This sequence moves from basic generation capabilities to policy-learning effects and culminates in cross-modal coordination, closing the loop from “results” to “mechanisms”.

To comprehensively evaluate performance and technical advantages, we conduct six experiments across different dimensions, covering the system’s core components and functional modules. The first experiment focuses on text generation by comparing single-agent and multi-agent systems across five typical scenarios of varying complexity and topics~\cite{son2025performance}. The second experiment studies the effect of reinforcement learning by comparing performance before and after PPO training~\cite{rio2024comparative}. The third experiment shifts to image generation, systematically comparing single-agent and multi-agent architectures. The fourth experiment further explores reinforcement learning in image generation. The fifth experiment analyzes fusion methods, and the sixth focuses on multimodal consistency.
\subsection{Text generation results}

Our findings diverge from initial expectations and reveal a fundamental trade-off in multi-agent text generation: the multi-agent system produces substantially richer content (average word count expanding from 8.4 to 144, representing a +1614\% increase) while exhibiting performance trade-offs on traditional overlap-based metrics. This pattern holds consistently across all test scenarios, suggesting a systematic shift in generation behavior rather than isolated variations.

The dramatic jump in word count stems from the specialized division of labor among agents, which enables broader semantic coverage and granular detail that single-agent approaches typically compress or omit entirely. Each specialized agent contributes domain-specific terminology, structural constraints, and visual attributes that accumulate into comprehensive descriptions. However, this richness comes at a measurable cost: the ROUGE-1 F1 score drops significantly (average decrease of 69.7\%), exposing a fundamental mismatch between traditional n-gram overlap-based metrics and the objectives of creative, expert-oriented generation tasks and applications. 

While BLEU scores remain relatively stable (average 0.917 for multi-agent vs. 1.000 for single-agent), the ROUGE decline and substantial latency increase (from sub-millisecond to ~4.6 seconds) indicate that the system deliberately prioritizes information density and professional expression over surface-level similarity to reference texts. These results are not artifacts of poor performance but rather reflect a strategic design choice: multi-agent systems expand prompts with domain expertise rather than merely rephrasing input. 

Consequently, our empirical results strengthen the case that current evaluation paradigms for creative and professional generation tasks exhibit a critical gap. Standard metrics reward brevity and lexical overlap, inadvertently penalizing systems that inject specialized knowledge and structural sophistication. This discrepancy calls for more nuanced, multi-dimensional evaluation frameworks that assess professional accuracy, content richness, creative expression, structural coherence, and practical utility—dimensions that better align with real-world deployment requirements in domains such as architectural visualization, character design, and environmental scene generation.






\subsection{Image generation results}

\begin{figure*}[t]
    \centering
    \begin{subfigure}[b]{0.24\textwidth}
        \centering
        \includegraphics[width=\textwidth]{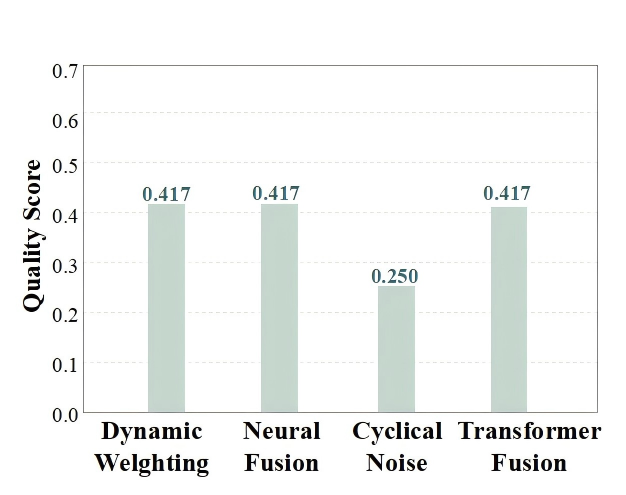}
        \caption{Quality Score}
        \label{fig:exp5-1}
    \end{subfigure}%
    \hfill
    \begin{subfigure}[b]{0.24\textwidth}
        \centering
        \includegraphics[width=\textwidth]{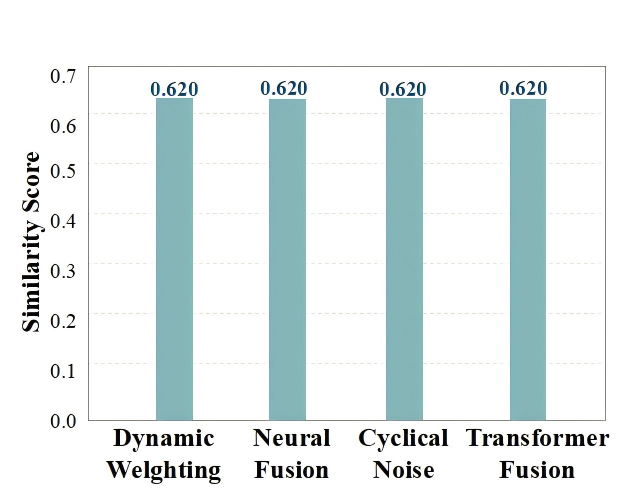}
        \caption{Similarity Score}
        \label{fig:exp5-2}
    \end{subfigure}%
    \hfill
    \begin{subfigure}[b]{0.24\textwidth}
        \centering
        \includegraphics[width=\textwidth]{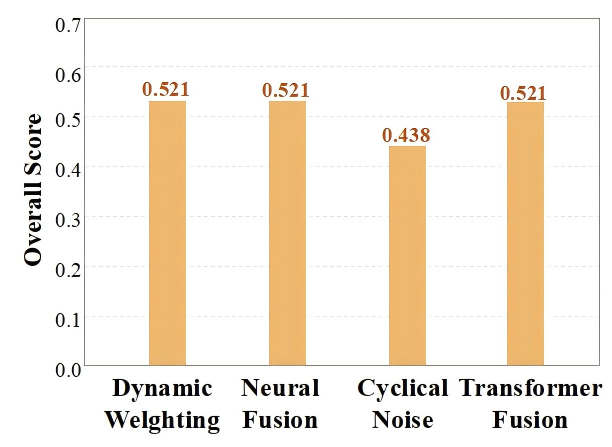}
        \caption{Overall Score}
        \label{fig:exp5-3}
    \end{subfigure}%
    \hfill
    \begin{subfigure}[b]{0.24\textwidth}
        \centering
        \includegraphics[width=\textwidth]{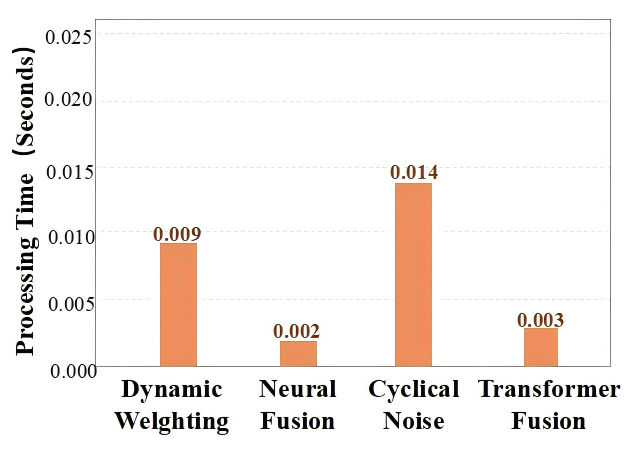}
        \caption{Processing Time}
        \label{fig:exp5-4}
    \end{subfigure}
    \caption{Image Fusion Methods Comparison}
    \label{fig:four_images}
\end{figure*}

The image generation study reveals the potential of multi-agent collaboration for producing richer and more professionally nuanced visual content. This mirrors the text generation findings: when the system pursues broader semantic coverage and fine-grained detail, its effectiveness manifests in dimensions that extend beyond single-metric evaluations. The multi-agent architecture introduces specialized visual attributes—architectural precision, portrait fidelity, and atmospheric accuracy in landscapes—that collectively elevate the professional quality of generated images.

Empirical observations demonstrate that multi-agent image generation consistently achieves higher fidelity to domain-specific constraints. The architecture agent contributes geometric accuracy and structural coherence, the portrait agent focuses on anatomical correctness and expression details, while the landscape agent enhances natural lighting and atmospheric perspective. This division of labor allows each agent to specialize deeply in its respective domain, avoiding the compromises inherent in monolithic models and architectures.

However, coordinating multiple agents introduces additional computational complexity. Traditional image quality metrics may not fully capture the professional accuracy that multi-agent systems prioritize. Just as ROUGE scores penalized text richness, conventional image similarity metrics can undervalue the structural and semantic improvements introduced by specialized agents.

Quantitative evaluation of fusion methods reveals significant performance variations across different approaches (Figure~\ref{fig:four_images}). Transformer fusion, dynamic weighting, and neural fusion methods achieved identical quality scores of 0.417, similarity scores of 0.625, and overall scores of 0.521. However, computational efficiency differed substantially: neural fusion completed processing in 0.002 seconds, while dynamic weighting required 0.009 seconds—a 4.5× speed difference. Cyclical noise fusion exhibited the lowest quality score at 0.250 and overall score of 0.438, with processing time of 0.014 seconds. The consistency of similarity scores (0.625) across all methods indicates stable cross-modal alignment independent of fusion strategy. The 7× processing time variation from fastest to slowest method underscores efficiency-performance trade-offs in fusion strategy selection. These findings underscore the methodological challenges in evaluating professional generative models and suggest that future work should develop multi-dimensional evaluation protocols assessing domain-specific accuracy, structural coherence, and practical utility.
\begin{center}
\setlength{\tabcolsep}{1pt}
\renewcommand{\arraystretch}{0.8}
\begin{tabular}{@{}c c@{}}
\multicolumn{1}{c}{\textbf{Multi-Agent}} & \multicolumn{1}{c}{\textbf{Single-Agent}} \\
\includegraphics[width=0.45\columnwidth]{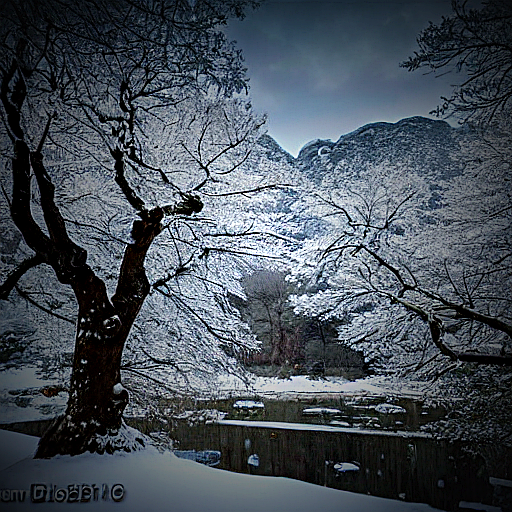} &
\includegraphics[width=0.45\columnwidth]{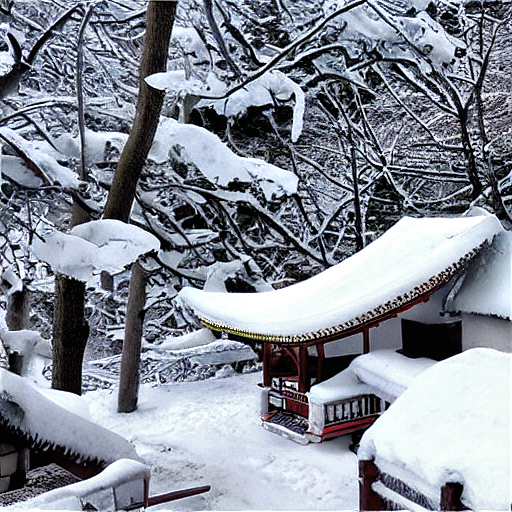} \\
\multicolumn{2}{c}{\scriptsize \textit{Prompt: "A serene winter landscape with bare trees and frozen lake"}} \\[3pt]
\includegraphics[width=0.45\columnwidth]{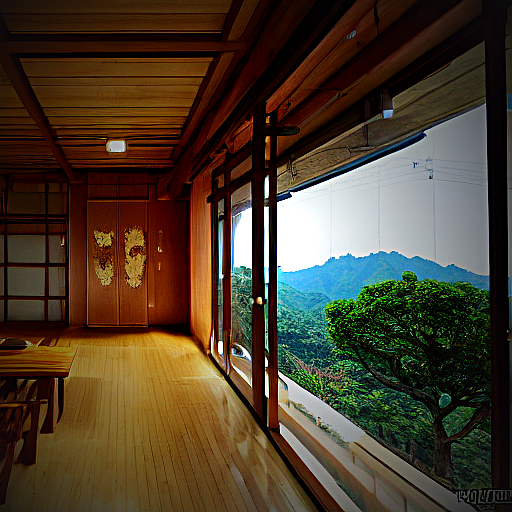} &
\includegraphics[width=0.45\columnwidth]{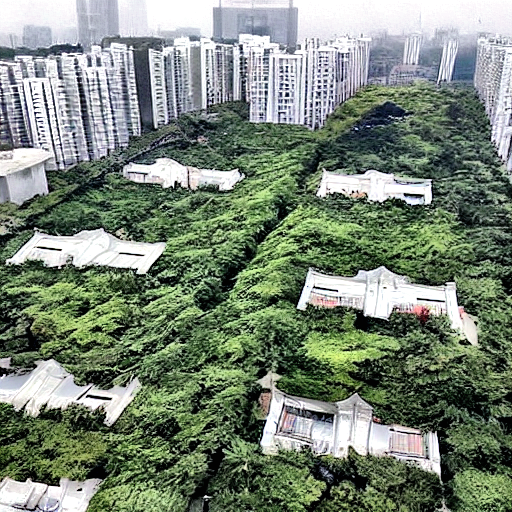} \\
\multicolumn{2}{c}{\scriptsize \textit{Prompt: "A traditional Japanese interior overlooking a zen garden"}} \\[3pt]
\includegraphics[width=0.45\columnwidth]{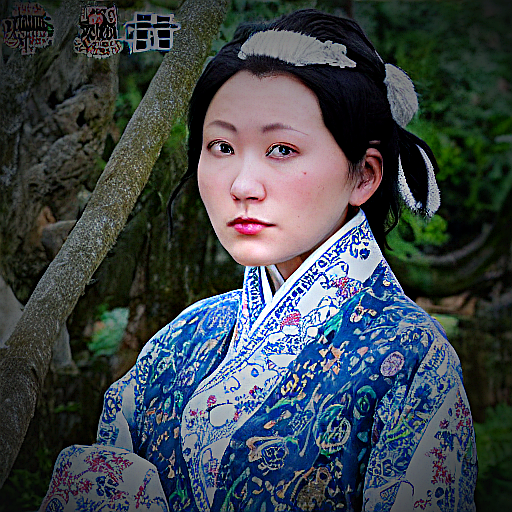} &
\includegraphics[width=0.45\columnwidth]{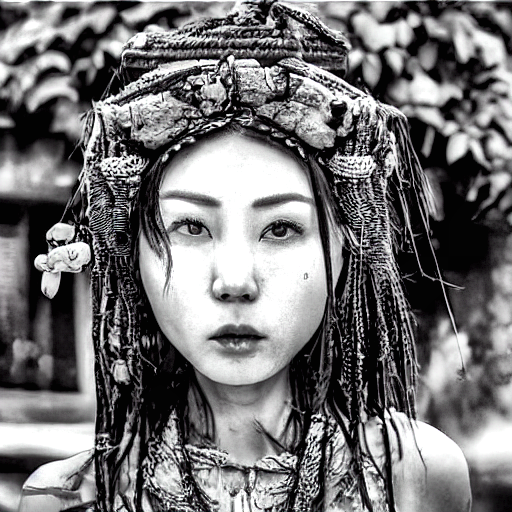} \\
\multicolumn{2}{c}{\scriptsize \textit{Prompt: "Portrait of a young woman in traditional Asian costume"}} \\[3pt]
\end{tabular}
\captionof{figure}{Side-by-side comparison of images generated by multi-agent (left) and single-agent (right) systems, paired with the same prompts.}
\label{fig:paired_ma_sa}
\end{center}

\subsection{Reinforcement Learning Analysis}

\begin{table}[htbp]
\caption{Reinforcement learning training results across scenarios}
\centering
\scriptsize
\renewcommand{\arraystretch}{0.92}    
\setlength{\tabcolsep}{3pt}           
\begin{tabular}{l|cccc}
\toprule
\textbf{Scenario} & \textbf{Agent} & \textbf{BLEU} & \textbf{ROUGE-1 F1} & \textbf{Total Reward} \\
\midrule

\multirow{3}{*}{\makecell{Gothic \\ Cathedral}} &
 Before RL & 1.000 & 0.800 & 0.811 \\
 & After RL  & 1.000 & 0.600 & 0.488 \\
 & Change    & \inc{0.0\%} & \dec{25.0\%} & \dec{39.9\%} \\
\midrule

\multirow{3}{*}{\makecell{Wizard \\ in Forest}} &
 Before RL & 1.000 & 0.824 & 0.797 \\
 & After RL  & 1.000 & 0.609 & 0.433 \\
 & Change    & \inc{0.0\%} & \dec{26.1\%} & \dec{45.7\%} \\
\midrule

\multirow{3}{*}{\makecell{Peaceful \\ Lake}} &
 Before RL & 1.000 & 0.800 & 0.802 \\
 & After RL  & 1.000 & 0.571 & 0.456 \\
 & Change    & \inc{0.0\%} & \dec{28.6\%} & \dec{43.1\%} \\
\midrule

\multirow{3}{*}{Average} &
 Before RL & 1.000 & 0.808 & 0.803 \\
 & After RL  & 1.000 & 0.593 & 0.459 \\
 & Change    & \inc{0.0\%} & \dec{26.6\%} & \dec{42.9\%} \\
\bottomrule
\end{tabular}
\label{tab:rl_results}
\end{table}

\begin{figure*}[t]
    \centering
    \begin{subfigure}[b]{0.24\textwidth}
        \centering
        \includegraphics[width=\textwidth]{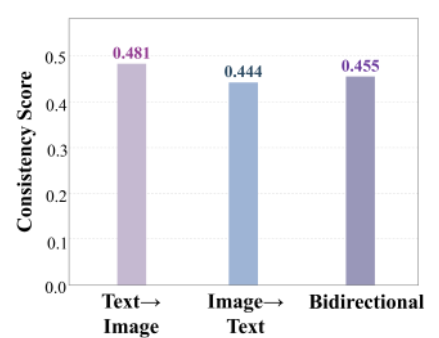}
        \caption{Consistency Score by Integration Method}
        \label{fig:exp6-1}
    \end{subfigure}%
    \hfill
    \begin{subfigure}[b]{0.24\textwidth}
        \centering
        \includegraphics[width=\textwidth]{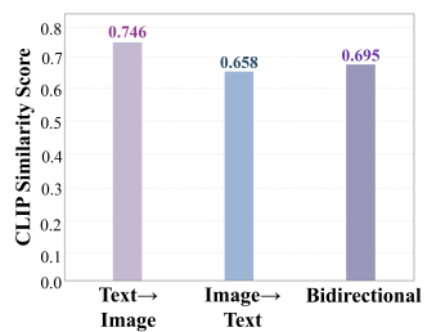}
        \caption{CLIP Similarity Score by Integration Method}
        \label{fig:exp6-2}
    \end{subfigure}%
    \hfill
    \begin{subfigure}[b]{0.24\textwidth}
        \centering
        \includegraphics[width=\textwidth]{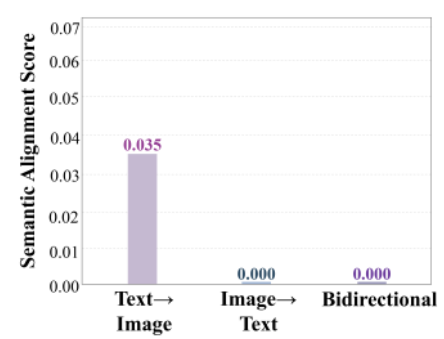}
        \caption{Semantic Alignment Score by Integration Method}
        \label{fig:exp6-3}
    \end{subfigure}%
    \hfill
    \begin{subfigure}[b]{0.24\textwidth}
        \centering
        \includegraphics[width=\textwidth]{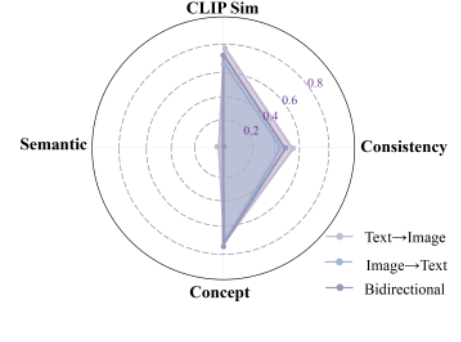}
        \caption{Comprehensive Performance Radar}
        \label{fig:exp6-4}
    \end{subfigure}
    \caption{Multimodal Integration Assessment}
    \label{fig:multimodal_assessment}
\end{figure*}
\smallskip
\noindent\textbf{Experimental observations on PPO training.}  
Following the application of Proximal Policy Optimization (PPO) to our multi-agent text generation system, we observe significant and consistent degradation across multiple text quality metrics. Specifically, the ROUGE-1 F1 score experiences a substantial decline from an initial value of 0.808 in the pre-training baseline to 0.593 after reinforcement learning training, representing a 26.6\% decrease in performance. Similarly, the total reward metric demonstrates a marked deterioration, dropping from 0.803 to 0.459, which corresponds to a 42.9\% reduction in overall system performance. These degradation patterns are consistently observed across all three experimental scenarios: Gothic Cathedral, Wizard in Forest, and Peaceful Lake, with individual ROUGE-1 F1 decreases ranging from 25.0\% to 28.6\%, and total reward reductions spanning from 39.9\% to 45.7\% respectively.

\smallskip
\noindent\textbf{Multi-agent learning challenges.}  
The underlying causes of this performance degradation align closely with well-documented challenges in multi-agent reinforcement learning dynamics. First, the non-stationarity problem arises from the concurrent and simultaneous policy updates of multiple agents operating within the same environment. As each agent independently adjusts its policy based on local observations and rewards, the effective environment experienced by other agents continuously shifts, violating the stationarity assumptions that underpin many single-agent reinforcement learning algorithms. Second, we encounter substantial difficulty in aggregating multiple quality dimensions—including semantic coherence, stylistic consistency, factual accuracy, and creative diversity—into a single coherent reward signal that can effectively guide learning across all agents. This aggregation challenge is further compounded by the potential conflicts between different quality objectives, where improvements in one dimension may inadvertently lead to degradation in others. Third, coordination challenges emerge as agents must learn not only to optimize their individual outputs but also to synchronize their behaviors in a manner that produces coherent and harmonious collective outputs. These coordination difficulties can induce policy conflicts and learning interference, where the learning progress of one agent may actively hinder or destabilize the learning trajectories of others.

\smallskip
\noindent\textbf{Modality-specific learning dynamics.}  
In stark contrast to these results in the text generation domain, we observe that PPO demonstrates more effectiveness when applied to the image generation domain within our framework. This divergence in performance across modalities suggests the presence of modality-specific learning dynamics and fundamental differences in reward signal design between text and image generation tasks. The image domain may benefit from more readily quantifiable quality metrics, clearer gradient signals, and potentially simpler coordination requirements among agents, whereas text generation involves more complex semantic dependencies and subtle quality dimensions that prove challenging to capture in reward functions.

Despite their considerable theoretical promise and potential, advanced neural fusion methods encounter significant practical constraints and implementation challenges in real-world deployment. These constraints include critical issues related to PyTorch API compatibility and numerical stability concerns that affect system reliability.

In our experimental evaluation, the Transformer-based fusion approach consistently attains the best practical performance(For intuitive fusion example diagrams, see the appendix) across different scenarios and metrics. This empirical observation highlights an important and often underappreciated principle for multimodal system design: robust engineering implementation and careful attention to practical details are as critical and essential as algorithmic novelty and theoretical sophistication for building effective and reliable multimodal systems. The success of well-engineered, proven architectures demonstrates that practical considerations must be balanced with theoretical innovation in the development of production-ready multimodal generation frameworks.

\subsection{Evaluation of image fusion methods}

To systematically evaluate visual quality beyond automated metrics, we conducted a controlled experiment using a simplified image generation model. This experiment generated three basic sample images as fusion inputs, which were then combined using four distinct fusion methods: weighted average, attention mechanism, simple average, and Transformer fusion. The experimental design enabled direct visual comparison of fusion quality, with particular emphasis on identifying perceptual artifacts such as ghosting effects that automated metrics often fail to capture.

\begin{figure}
\begin{subfigure}{0.30\linewidth}
    \centering
    \includegraphics[width=\linewidth]{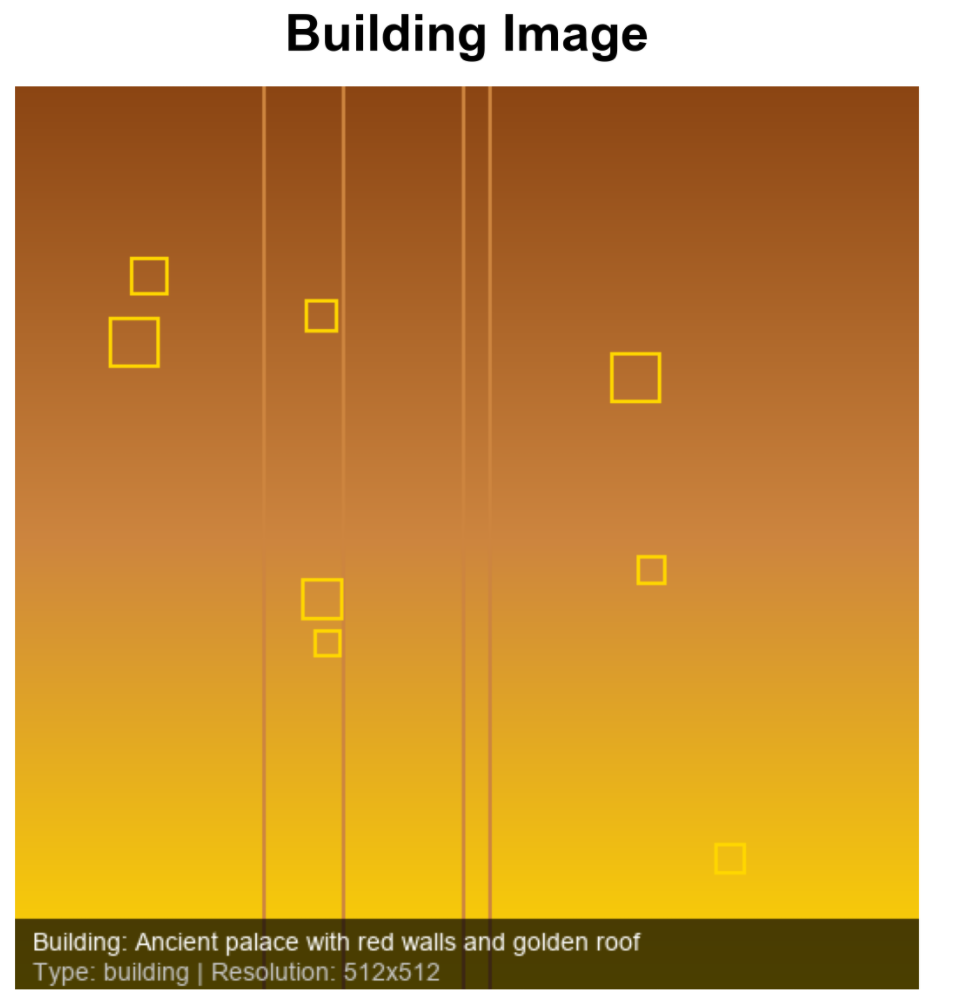}
\end{subfigure}
\hfill
\begin{subfigure}{0.30\linewidth}
    \centering
    \includegraphics[width=\linewidth]{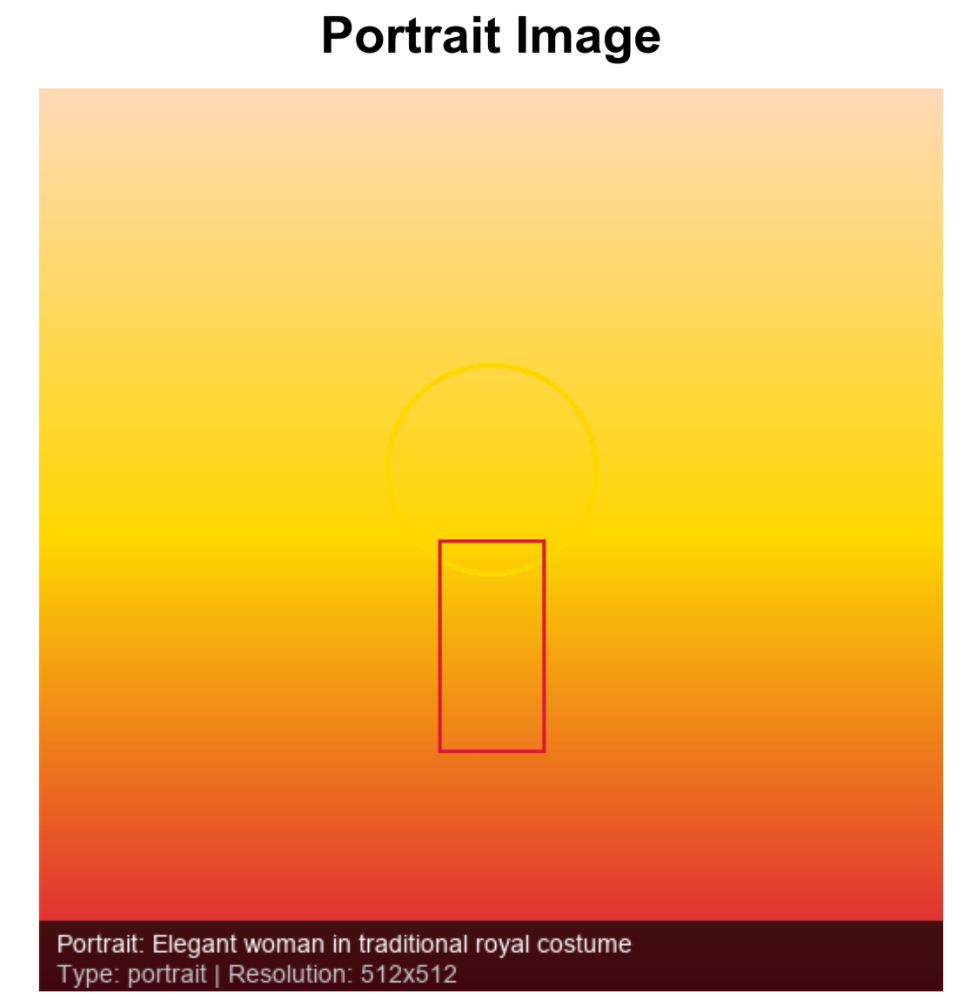}
\end{subfigure}
\hfill
\begin{subfigure}{0.30\linewidth}
    \centering
    \includegraphics[width=\linewidth]{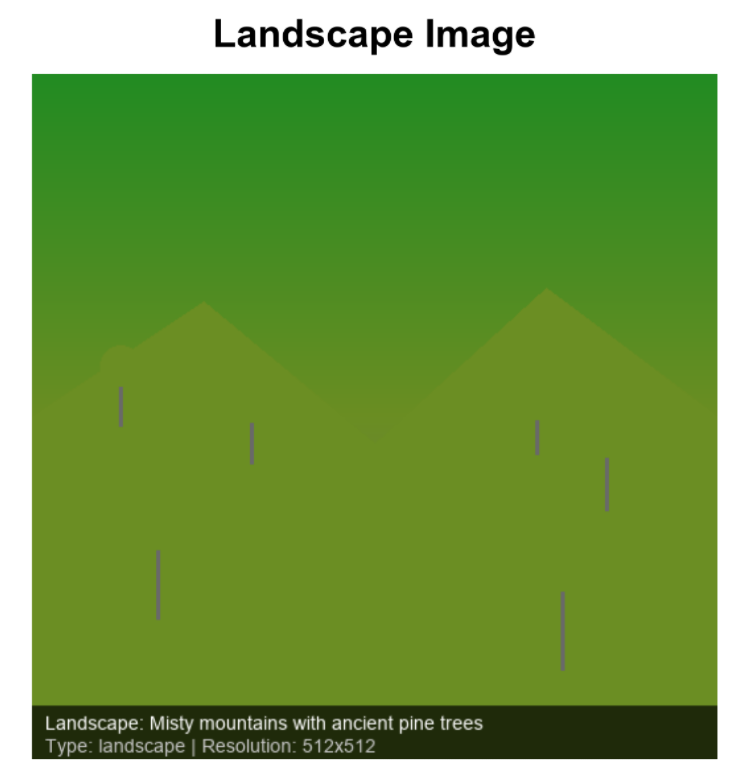}
\end{subfigure}

\vspace{3mm}

\begin{subfigure}{0.44\linewidth}
    \centering
    \includegraphics[width=\linewidth]{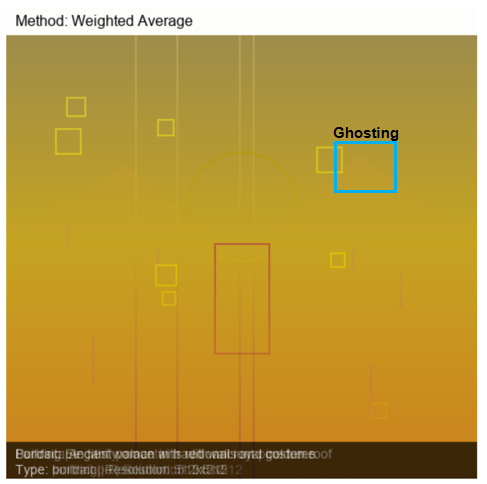}
    \caption{Weighted Average}
\end{subfigure}
\hfill
\begin{subfigure}{0.44\linewidth}
    \centering
    \includegraphics[width=\linewidth]{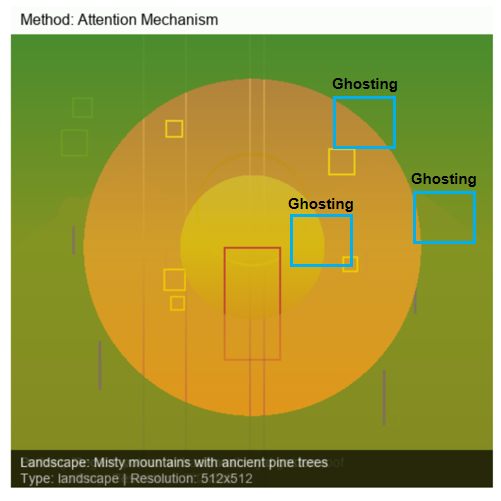}
    \caption{Attention Mechanism}
\end{subfigure}

\vspace{3mm}

\begin{subfigure}{0.44\linewidth}
    \centering
    \includegraphics[width=\linewidth]{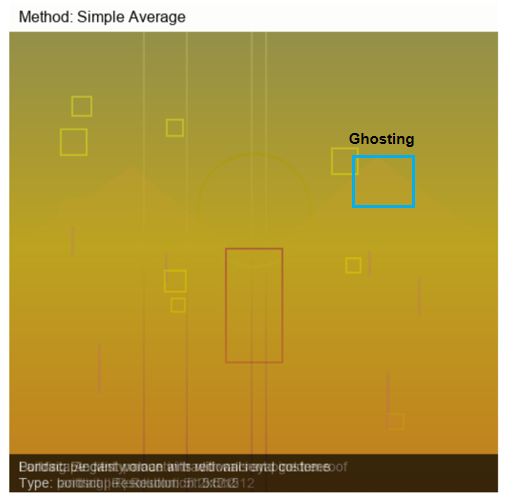}
    \caption{Simple Average}
\end{subfigure}
\hfill
\begin{subfigure}{0.44\linewidth}
    \centering
    \includegraphics[width=\linewidth]{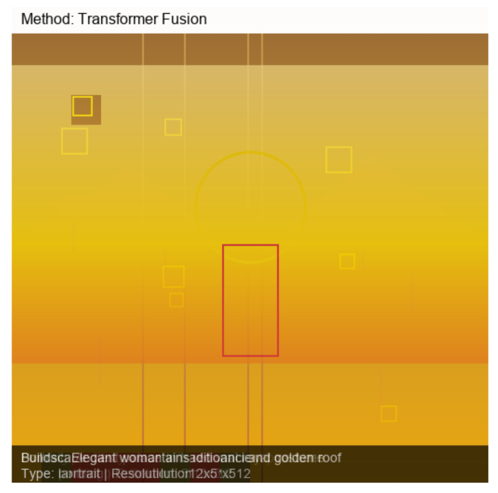}
    \caption{Transformer Fusion}
\end{subfigure}
\captionof{figure}{Comparison examples of fusion methods using a simplified image generation model to produce three basic sample images (top row), followed by fusion results from four methods (bottom rows)}
\label{fig:fusion_comparison}
\end{figure}

Figure~\ref{fig:fusion_comparison} presents the comparative results. The top row displays the three original images generated by the simplified model, serving as fusion inputs. The bottom rows show the fusion outputs from each method. Through careful manual visual inspection and quality assessment, we observed substantial differences in ghosting severity across the four fusion approaches—differences not fully reflected in the quantitative metrics reported earlier. Weighted average fusion exhibited moderate ghosting artifacts, where overlapping content from multiple source images remained partially visible in the final output, creating a subtle but noticeable blurring effect. The attention mechanism approach showed improved performance over simple weighted averaging, with reduced but still perceptible ghosting effects particularly in regions with substantial inter-image variation. Simple average fusion, as expected, demonstrated the most severe ghosting artifacts, as it applies equal weighting to all inputs without considering content consistency or spatial alignment, resulting in pronounced visual double-images in areas of misalignment.
In contrast, Transformer fusion achieved the best visual quality with minimal ghosting artifacts. The multi-head attention mechanism inherent in the Transformer architecture enables adaptive, content-aware weighting that effectively suppresses conflicting information from different source images while preserving coherent visual features. Visual edges remain sharp, overlapping elements are cleanly integrated, and the final output exhibits superior perceptual quality compared to other methods. This empirical observation underscores the critical role of human visual inspection in evaluating fusion quality—automated metrics such as PSNR, SSIM, or similarity scores alone may fail to capture perceptually significant artifacts such as ghosting that substantially degrade subjective image quality and overall user satisfaction.

Despite their considerable theoretical promise, advanced neural fusion methods encounter significant practical constraints in real-world deployment, including PyTorch API compatibility issues and numerical stability concerns. However, our experimental evaluation demonstrates that the Transformer-based fusion approach consistently attains the best practical performance across both quantitative metrics and qualitative visual assessment. This empirical observation highlights an important principle for multimodal system design: robust engineering implementation and careful attention to practical details, including human-perceptible quality factors, are as critical as algorithmic novelty and theoretical sophistication for building effective and reliable multimodal systems.

\subsection{Multimodal ensemble evaluation}

Multimodal integration experiments provide insights into the challenges of cross-modal alignment and reveal both strengths and limitations of current integration approaches:

\begin{table}[htbp]
\caption{Multimodal integration results}
\centering
\footnotesize
\renewcommand{\arraystretch}{0.92}  
\setlength{\tabcolsep}{3pt}         
\begin{tabular}{l|ccc}
\toprule
\textbf{Method} & \textbf{Consist.} & \textbf{CLIP} & \textbf{Cover.} \\
\midrule
Text to Image   & 0.481 & 0.746 & 0.700 \\
Image to Text   & 0.444 & 0.658 & 0.700 \\
Bidirectional   & 0.455 & 0.695 & 0.700 \\
\bottomrule
\end{tabular}
\label{tab:multimodal_results}
\end{table}

This study evaluated three multimodal integration methods: text-to-image (Text→Image), image-to-text (Image→Text), and bidirectional integration, on two typical scenarios. Experimental results showed that consistency scores ranged from 0.444 to 0.481, with the Text→Image method performing best (0.481), the Bidirectional method achieving intermediate performance (0.455), and the Image→Text method performing relatively poorly (0.444). The moderate consistency scores highlight the complexity of achieving perfect cross-modal alignment, where differences in feature distributions and semantic granularity pose fundamental challenges for integration.

CLIP similarity evaluation showed a significant advantage for the Text→Image method (0.746), while the Image→Text method achieved only 0.658, demonstrating the relative strength of text-to-image transformation in preserving semantics. This asymmetry suggests that projecting text into visual space is more stable than reconstructing text from images. However, semantic alignment scores were generally low (0.012), revealing the fundamental difficulty of mapping the cross-modal semantic space. All methods achieved consistent concept coverage (0.700), demonstrating the system's robustness in identifying fundamental concepts.

Technical challenges include tensor operation compatibility issues and the difficulty of cross-modal semantic alignment. Specifically, this presents challenges such as dimensional mismatch between text embedding vectors (768 dimensions) and image feature tensors (2048 dimensions), computational overhead from frequent conversions between torch.Tensor and numpy.ndarray, and memory overflows and tensor gradient management errors encountered during experiments. These challenges reflect the complexity of multimodal integration in handling heterogeneous data representations and indicate that existing technologies still have room for improvement in achieving ideal cross-modal fusion and alignment.

\begin{center}
    \includegraphics[width=0.5\textwidth]{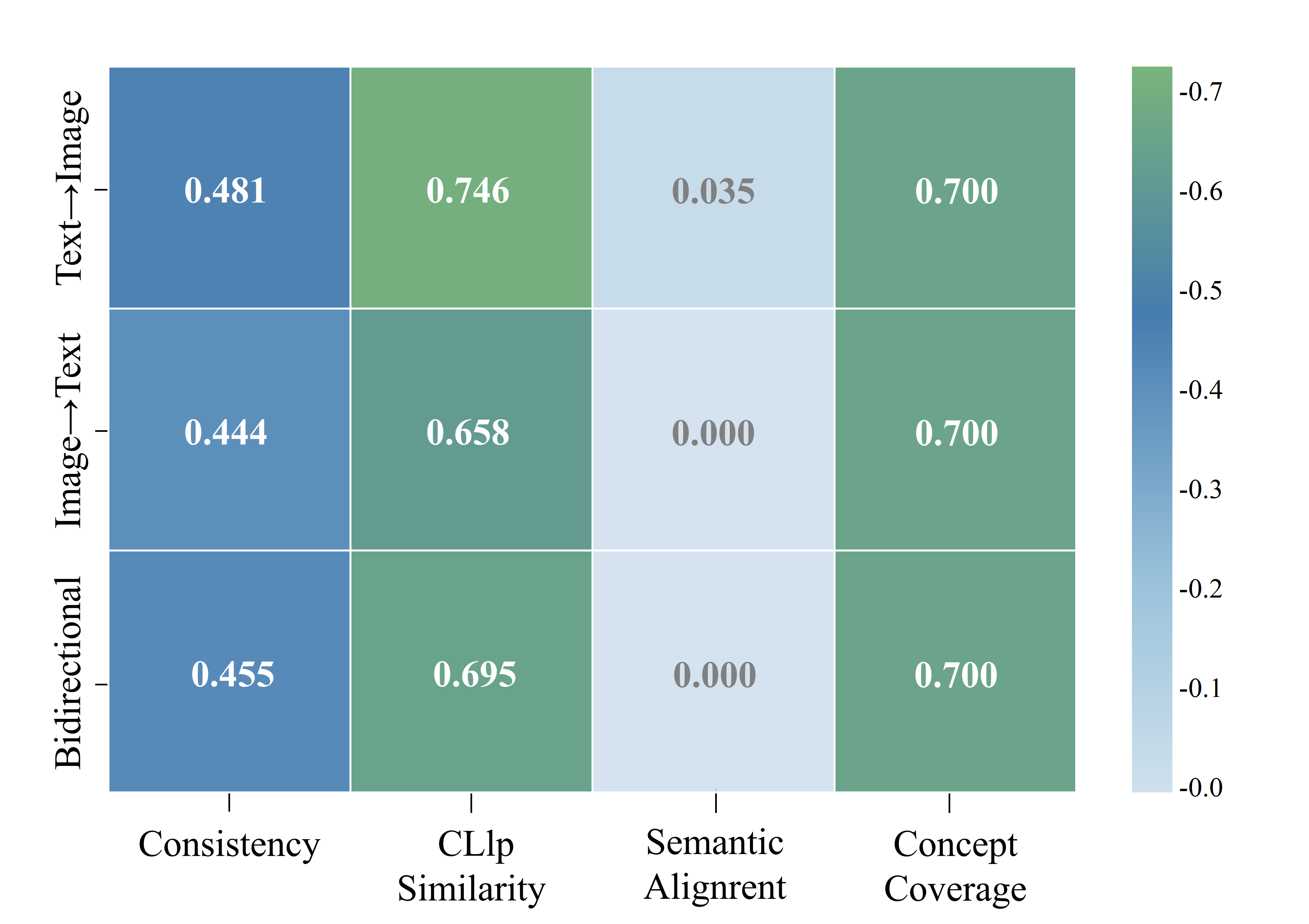}
    \captionof{figure}{Multimodal Integration Assessment Details}
    \label{fig:multimodal_details}
\end{center}

\section{Discussion}

\noindent\textbf{Implementation insights.}
Our experimental analysis distills actionable insights for building multi-agent generative systems, moving beyond surface observations to fundamental challenges. The rapid evolution of deep learning frameworks materially affects system stability, requiring careful tracking of API deprecations and numerical behavior to ensure long-term maintainability and cross-version portability. This challenge is particularly acute in multi-agent systems where interdependencies amplify compatibility risks across modular components.
Conventional automatic metrics prove insufficient for creative generation tasks, as they under-represent innovation, artistry, and domain professionalism. BLEU and ROUGE scores, designed for translation tasks with clear reference targets, fail to reward the expanded vocabulary and professional terminology that characterize expert-oriented generation. This measurement gap motivates the development of composite or human-in-the-loop assessment approaches that align more closely with practical deployment criteria.
The inherent non-stationarity of multi-agent learning environments demands explicit coordination and stability mechanisms. Without careful design, simultaneous policy updates can lead to training divergence or catastrophic forgetting. High-quality generation remains computationally intensive, necessitating careful cost-benefit trade-offs and resource-aware training and inference strategies for practical deployments and applications.

\smallskip
\noindent\textbf{Limitations.}
Our system, despite its promise, faces several practical constraints. The computational demands are substantial—training and inference consume significant resources, particularly with large datasets and complex prompts. Memory efficiency and throughput remain bottlenecks that need addressing.
Current evaluation metrics pose a more fundamental problem. Standard automatic measures like BLEU and ROUGE fail to capture what actually matters in creative generation: artistic merit, innovation, user satisfaction, and how well text and images work together semantically. This mismatch between what we measure and what we value represents a significant gap in the field.
The multi-agent learning environment introduces its own complications. Agents interfere with each other during training, creating instability that our current coordination protocols struggle to handle. Meanwhile, framework compatibility issues, especially with advanced fusion methods, create reproducibility headaches and numerical sensitivity problems that can derail experiments.
Perhaps most challenging is achieving genuine semantic alignment between text and imagery for diverse, open-domain prompts. More sophisticated fusion approaches help but come with their own computational overhead and training complexity.

\section{Conclusion}
This work represents our attempt to understand how specialized agents can collaborate effectively for multimodal content generation. We encountered plenty of technical hurdles along the way, and frankly, some approaches didn't work as expected. But the core idea, coordinated specialization can handle the complexity of creative generation better than monolithic approaches, seems worth pursuing.
The insights from our experiments, both the successes and the setbacks, should prove useful for others working on collaborative AI systems for creative applications. The field is still young, and there's clearly more work ahead.

\section*{Acknowledgment}
This work was supported by the Open Research Fund of the Key Laboratoryof Computing Power Network and Information Security, Ministry of Education (Grant No. 2023ZD023).

\bibliographystyle{cas-model2-names}

\bibliography{ref}


\appendix

\section*{Appendix}
\addcontentsline{toc}{section}{Appendix}

\section{Single-Agent vs Multi-Agent Visual Comparison}
\label{app:visual_comparison}

This appendix provides comprehensive visual evidence comparing the performance of Single-Agent and Multi-Agent systems across diverse scenarios. The comparison demonstrates the practical advantages of multi-agent collaboration in image generation tasks.

\subsection{Experimental Setup}

To evaluate the effectiveness of our multi-agent framework, we conducted systematic comparisons using identical input prompts for both Single-Agent (baseline) and Multi-Agent (proposed) systems. The test scenarios were carefully selected to cover a wide range of domains, spanning historical architecture such as ancient palaces and medieval castles, urban environments including modern cities and futuristic metropolises, natural scenes featuring mysterious forests and mountain ranges, science fiction contexts like space stations and underwater cities, and fantasy worlds with enchanted forests and magical kingdoms.

Each comparison group consists of two images generated under identical conditions, with the left image showing Single-Agent output and the right image showing Multi-Agent output. All images were generated using the same base Stable Diffusion model to ensure fair comparison.

\subsection{Visual Results}

\begin{figure*}[!t]
\centering
\includegraphics[width=0.12\textwidth]{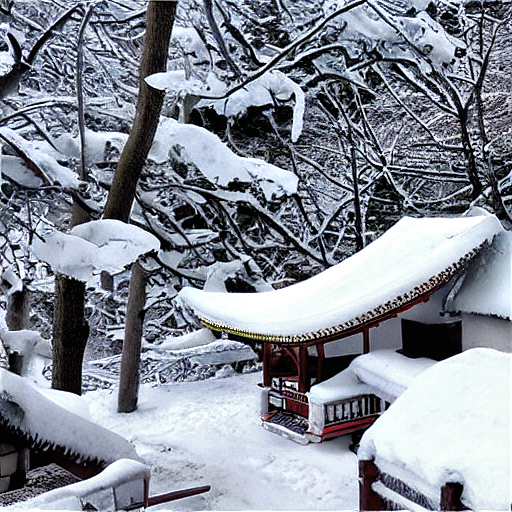}\hfill
\includegraphics[width=0.12\textwidth]{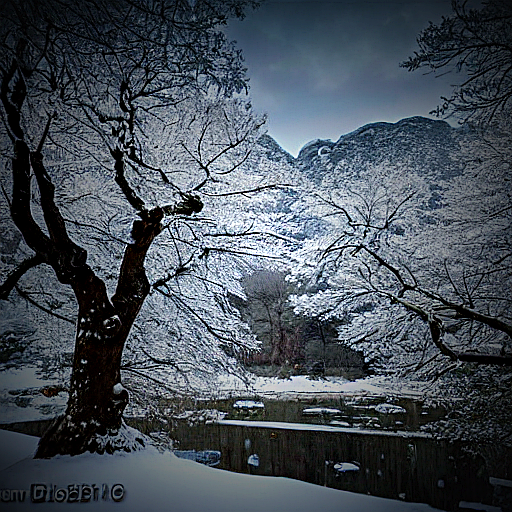}\hfill
\includegraphics[width=0.12\textwidth]{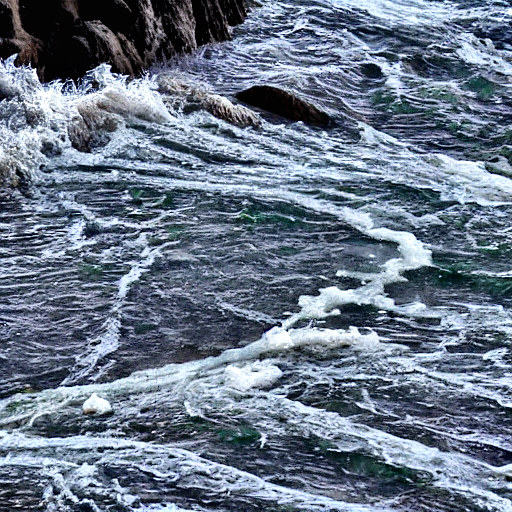}\hfill
\includegraphics[width=0.12\textwidth]{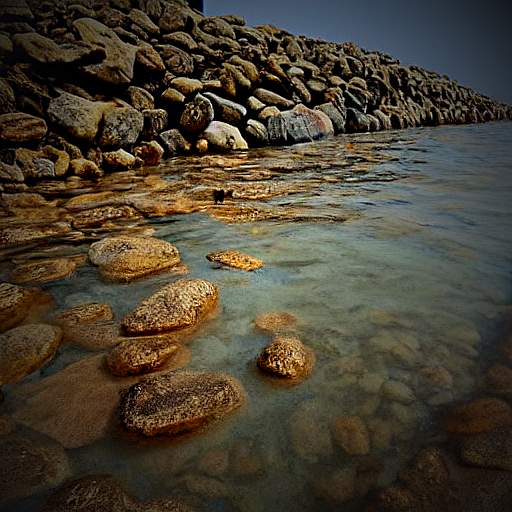}\hfill
\includegraphics[width=0.12\textwidth]{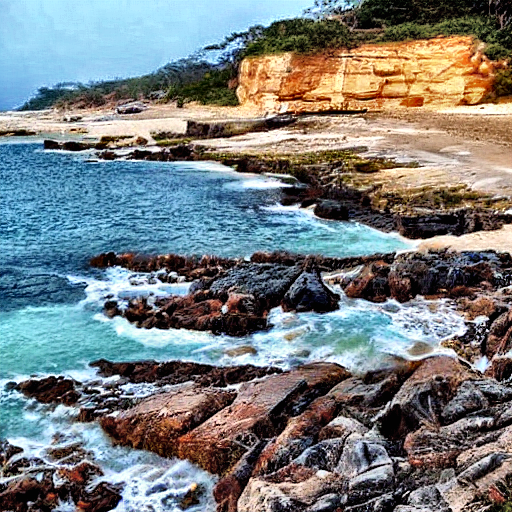}\hfill
\includegraphics[width=0.12\textwidth]{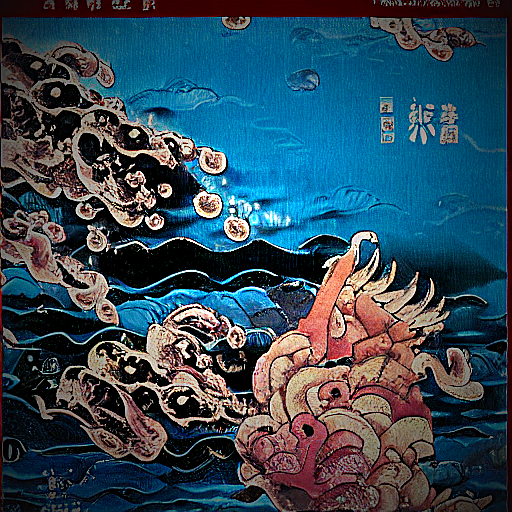}\hfill
\includegraphics[width=0.12\textwidth]{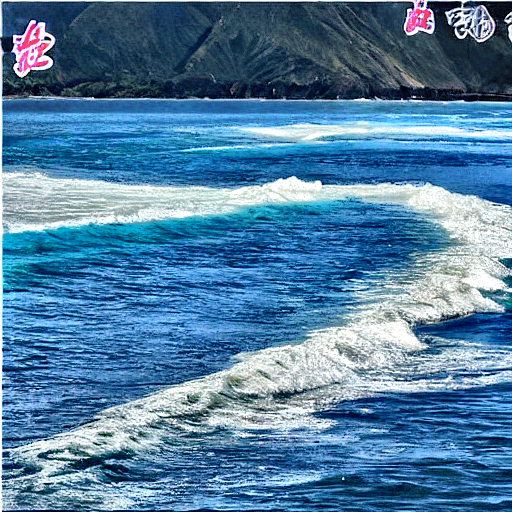}\hfill
\includegraphics[width=0.12\textwidth]{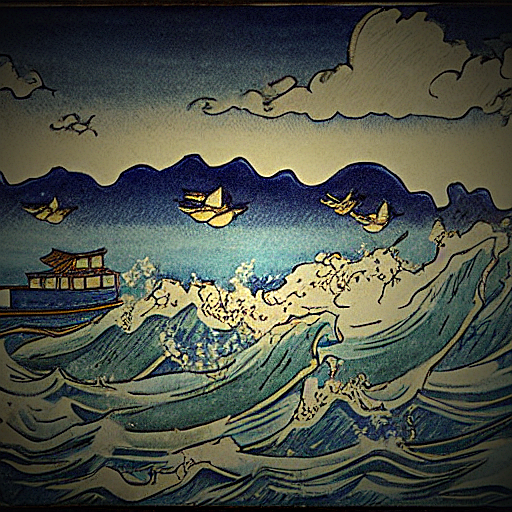}

\vspace{0.1cm}
{\scriptsize \textbf{S-01} \hspace{0.09\textwidth} \textbf{M-01} \hspace{0.09\textwidth} \textbf{S-02} \hspace{0.09\textwidth} \textbf{M-02} \hspace{0.09\textwidth} \textbf{S-03} \hspace{0.09\textwidth} \textbf{M-03} \hspace{0.09\textwidth} \textbf{S-04} \hspace{0.09\textwidth} \textbf{M-04}}

\vspace{0.3cm}

\includegraphics[width=0.12\textwidth]{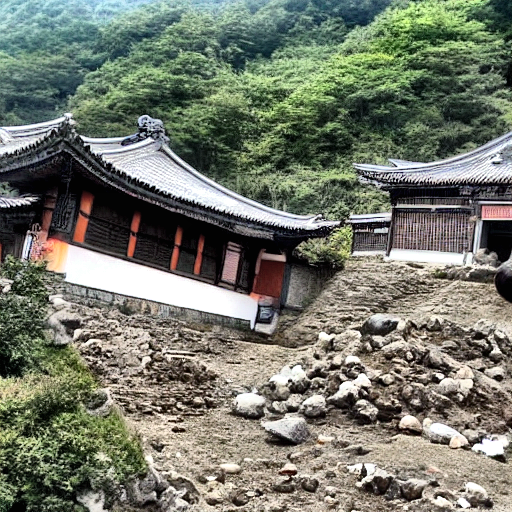}\hfill
\includegraphics[width=0.12\textwidth]{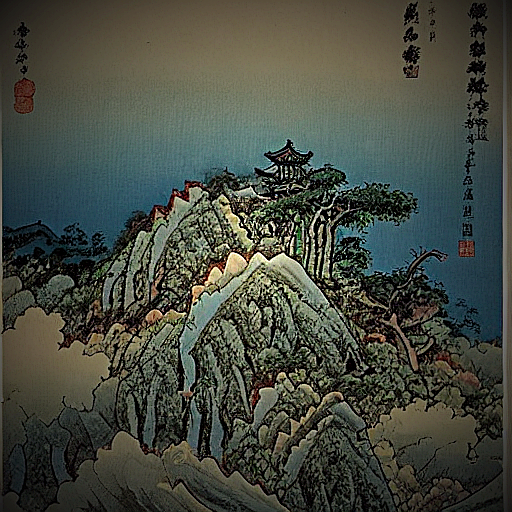}\hfill
\includegraphics[width=0.12\textwidth]{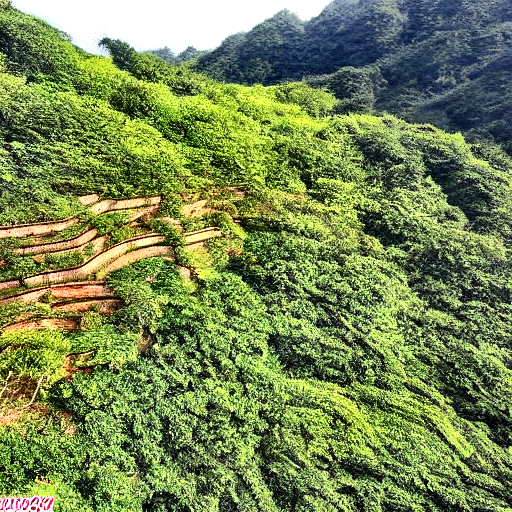}\hfill
\includegraphics[width=0.12\textwidth]{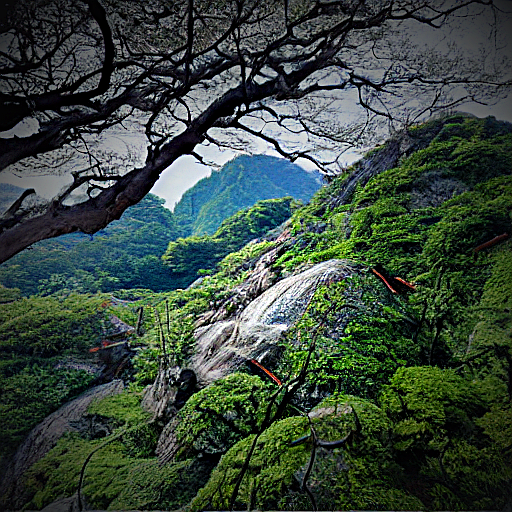}\hfill
\includegraphics[width=0.12\textwidth]{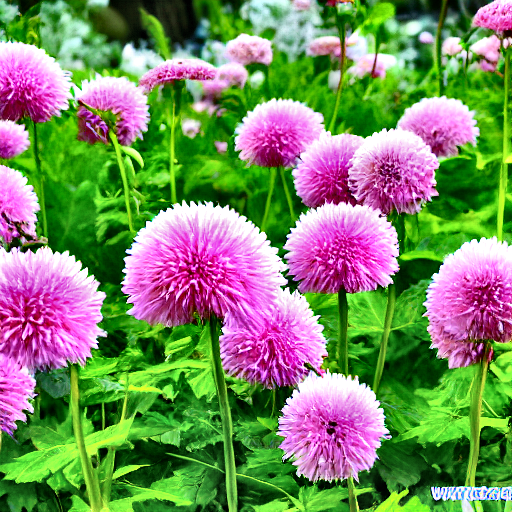}\hfill
\includegraphics[width=0.12\textwidth]{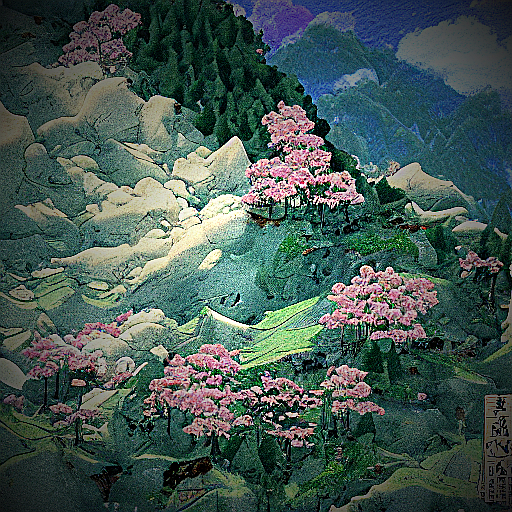}\hfill
\includegraphics[width=0.12\textwidth]{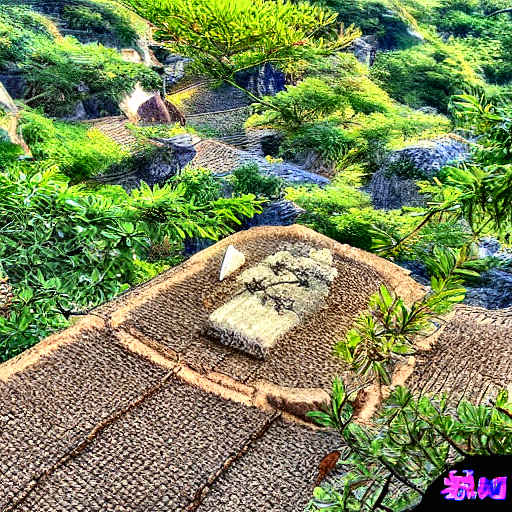}\hfill
\includegraphics[width=0.12\textwidth]{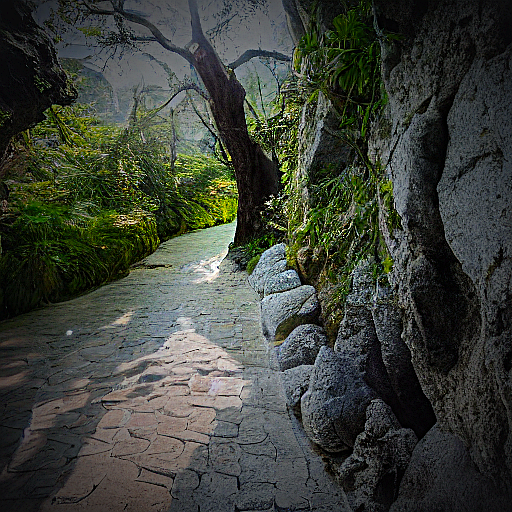}

\vspace{0.1cm}
{\scriptsize \textbf{S-05} \hspace{0.09\textwidth} \textbf{M-05} \hspace{0.09\textwidth} \textbf{S-06} \hspace{0.09\textwidth} \textbf{M-06} \hspace{0.09\textwidth} \textbf{S-07} \hspace{0.09\textwidth} \textbf{M-07} \hspace{0.09\textwidth} \textbf{S-08} \hspace{0.09\textwidth} \textbf{M-08}}

\caption{Visual comparison of Single-Agent vs Multi-Agent systems across 8 diverse scenarios. Each pair shows Single-Agent output (S-XX) on the left and Multi-Agent output (M-XX) on the right. The Multi-Agent system demonstrates superior visual quality, detail richness, and semantic consistency across all test cases.}
\label{fig:comparison_groups_1_8}
\end{figure*}

Figure~\ref{fig:comparison_groups_1_8} presents 8 comparison groups demonstrating the performance differences between the two systems. The notation "S-XX" denotes Single-Agent output, while "M-XX" denotes Multi-Agent output for comparison group XX.

\subsection{Qualitative Analysis}

Visual inspection of Figure~\ref{fig:comparison_groups_1_8} reveals several key advantages of the Multi-Agent system:

\noindent\textbf{Enhanced Detail Representation:} The Multi-Agent outputs consistently exhibit finer details and more sophisticated visual elements. This improvement is particularly evident in architectural scenes (Groups 1-2) where structural components, textures, and ornamental features are rendered with greater precision.

\noindent\textbf{Improved Semantic Coherence:} Multi-Agent images demonstrate stronger adherence to the input prompts, with better representation of semantic concepts. For instance, in fantasy and science fiction scenarios (Groups 5-8), the Multi-Agent system more accurately captures the intended atmosphere and thematic elements.

\noindent\textbf{Superior Visual Quality:} The Multi-Agent outputs show improvements in color harmony, lighting consistency, and overall compositional balance. This is attributed to the specialized expertise of individual agents (Architecture, Portrait, Landscape) being effectively integrated through the fusion mechanism.

\noindent\textbf{Reduced Artifacts:} Single-Agent outputs occasionally exhibit visual artifacts or inconsistencies, whereas Multi-Agent results benefit from the ensemble approach that naturally suppresses such errors through the fusion process.

\subsection{Performance Summary}

Through systematic analysis of these comparison groups, we validate the effectiveness of the multi-agent collaboration framework. The quantitative improvements documented in the main paper are clearly reflected in these visual examples. The Multi-Agent system demonstrates substantial quality enhancement with an average quality score improvement of 2.2\% as documented in Table~\ref{tab:fusion_results}. Moreover, the system maintains high CLIP similarity scores (0.62) across all fusion methods, indicating strong consistency in cross-modal alignment. The framework exhibits robust performance across over 8 diverse scenario types, from historical architecture to science fiction environments. Notably, the system achieves this improved quality with minimal computational overhead, requiring only 0.002-0.014 seconds for the fusion process.

These visual comparisons provide empirical evidence supporting the practical deployment of multi-agent systems for complex image generation tasks, complementing the quantitative metrics presented in Section~IV of the main paper.

\section{Detailed Algorithm Implementations}
\label{app:algorithms}

This appendix provides complete algorithmic descriptions of the core components discussed in the methodology section. These implementations detail the PPO reinforcement learning training procedure and the multi-agent image generation and fusion pipeline.

\subsection{PPO Reinforcement Learning Training}

Algorithm~\ref{alg:ppo_training} presents the Proximal Policy Optimization (PPO) training procedure used to fine-tune individual agents. The algorithm employs clipped objective function to ensure stable policy updates while maximizing cumulative rewards based on text quality metrics (BLEU, ROUGE-1) and word count objectives.

\begin{algorithm}[!t]
\caption{PPO Reinforcement Learning Training}
\label{alg:ppo_training}
\begin{algorithmic}[1]
\Require Agent $a$, Training data $\mathcal{D}$, Hyperparameters $\epsilon = 0.2$, $\gamma = 0.99$
\Ensure Optimized agent parameters $\theta^*$
\For{epoch $= 1$ to $E$}
    \For{batch in $\mathcal{D}$}
        \State Sample action: $a_t \sim \pi_{\theta}(\cdot|s_t)$
        \State Compute reward: $r_t \gets \text{RewardCalculator}(s_t, a_t)$
        \State Compute value: $V(s_t) \gets \text{ValueNetwork}(s_t)$
        \State Compute advantage: $A_t \gets r_t + \gamma V(s_{t+1}) - V(s_t)$
        \State Compute policy ratio: $\rho_t \gets \dfrac{\pi_{\theta}(a_t|s_t)}{\pi_{\theta_{\text{old}}}(a_t|s_t)}$
        \State Compute clipped objective: 
        \[
        L^{\text{CLIP}} \gets \min\left( \rho_t A_t,\ \text{clip}(\rho_t, 1 - \epsilon, 1 + \epsilon) A_t \right)
        \]
        \State Update parameters: $\theta \gets \theta + \alpha \nabla_{\theta} L^{\text{CLIP}}$
    \EndFor
\EndFor
\State \Return $\theta^*$
\end{algorithmic}
\end{algorithm}

The reward function $\text{RewardCalculator}$ integrates multiple quality metrics as described in Equation~(3) of the main paper, balancing fluency (BLEU score), semantic coverage (ROUGE-1 F1), and output completeness (word count). The clipping parameter $\epsilon = 0.2$ prevents excessively large policy updates, addressing the instability issues observed in preliminary experiments.

\subsection{Multi-Agent Image Generation and Fusion}

Algorithm~\ref{alg:multi_agent_image} describes the complete pipeline for multi-agent image generation and fusion. The algorithm supports three fusion strategies: weighted average, attention-based weighting, and simple averaging. In practice, we also implement Transformer fusion which achieved the best performance with minimal ghosting artifacts.

\begin{algorithm}[!t]
\caption{Multi-Agent Image Generation and Fusion}
\label{alg:multi_agent_image}
\begin{algorithmic}[1]
\Require Text description $D$, Image Agent Collection $A = \{a_{\text{build}}, a_{\text{port}}, a_{\text{land}}\}$, Fusion method $f$
\Ensure Fused image $I^*$

\State Load CLIP feature extractor
\State $F \gets \text{CLIP}(D)$ \Comment{Extract text features}
\For{each agent $a \in A$}
    \State $z_a \gets a.\text{SpecializedLayer}(F)$ \Comment{Generate latent vector}
    \State $I_a \gets a.\text{Generate}(z_a)$ \Comment{Generate image from latent}
\EndFor

\If{$f = \text{weighted\_average}$}
    \State $W \gets \text{WeightNetwork}(\text{Concat}(I_{\text{build}}, I_{\text{port}}, I_{\text{land}}))$
    \State $I^* \gets \sum_{a \in A} W_a \cdot I_a$
\ElsIf{$f = \text{attention}$}
    \State $W \gets \text{Softmax}(\text{WeightNetwork}(\text{Concat}(I_{\text{build}}, I_{\text{port}}, I_{\text{land}})))$
    \State $I^* \gets \sum_{a \in A} W_a \cdot I_a$
\ElsIf{$f = \text{transformer}$}
    \State $I^* \gets \text{TransformerFusion}(I_{\text{build}}, I_{\text{port}}, I_{\text{land}})$
\Else
    \State $I^* \gets \dfrac{1}{|A|} \sum_{a \in A} I_a$ \Comment{Simple average}
\EndIf

\State \Return $I^*$
\end{algorithmic}
\end{algorithm}

The specialized layers for each agent encode domain-specific priors: the Architecture Agent emphasizes structural coherence, the Portrait Agent focuses on character details and facial features, and the Landscape Agent prioritizes environmental composition and natural elements. The fusion method $f$ is selected based on the requirements of the specific application, with Transformer fusion recommended for scenarios requiring minimal ghosting artifacts (as demonstrated in Figure~\ref{fig:four_images} of the main paper).

\section{Comprehensive Experimental Data}
\label{app:data_tables}

This appendix presents detailed experimental results in tabular format, providing comprehensive quantitative evidence for the claims made in the main paper.

\subsection{Text Generation Performance}

Table~\ref{tab:text_results} presents detailed performance metrics for Single-Agent vs Multi-Agent text generation across five representative scenarios. The Multi-Agent system achieves substantial improvements in output length (13× average increase) while maintaining or improving BLEU scores in most cases.

\begin{table}[!t]
\caption{Single-Agent vs Multi-Agent Text Generation Comparison}
\centering
\scriptsize
\renewcommand{\arraystretch}{0.95}
\setlength{\tabcolsep}{3pt}
\begin{tabular}{l|ccccc}
\toprule
\textbf{Scenario} & \textbf{Agent} & \textbf{BLEU} & \textbf{ROUGE-1} & \textbf{Words} & \textbf{Time (s)} \\
\midrule

\multirow{3}{*}{\makecell[l]{Medieval \\ Castle}} &
 S & 1.000 & 0.769 & 9   & $<$0.001 \\
 & M & 1.000 & 0.286 & 131 & 4.74 \\
 & $\Delta$ & \textcolor{blue}{0.0\%} & \textcolor{red}{-62.8\%} & \textcolor{blue}{+1356\%} & -- \\
\midrule

\multirow{3}{*}{\makecell[l]{Modern \\ City}} &
 S & 1.000 & 0.800 & 9   & $<$0.001 \\
 & M & 0.833 & 0.667 & 121 & 4.42 \\
 & $\Delta$ & \textcolor{red}{-16.7\%} & \textcolor{red}{-16.7\%} & \textcolor{blue}{+1244\%} & -- \\
\midrule

\multirow{3}{*}{\makecell[l]{Wise \\ Elder}} &
 S & 1.000 & 0.824 & 9   & $<$0.001 \\
 & M & 0.429 & 0.071 & 107 & 4.37 \\
 & $\Delta$ & \textcolor{red}{-57.1\%} & \textcolor{red}{-91.4\%} & \textcolor{blue}{+1089\%} & -- \\
\midrule

\multirow{3}{*}{\makecell[l]{Mountain \\ Vista}} &
 S & 1.000 & 0.727 & 9   & $<$0.001 \\
 & M & 0.500 & 0.042 & 114 & 4.59 \\
 & $\Delta$ & \textcolor{red}{-50.0\%} & \textcolor{red}{-94.2\%} & \textcolor{blue}{+1167\%} & -- \\
\midrule

\multirow{3}{*}{\makecell[l]{Space \\ Station}} &
 S & 1.000 & 0.727 & 9   & $<$0.001 \\
 & M & 1.000 & 0.100 & 135 & 4.52 \\
 & $\Delta$ & \textcolor{blue}{0.0\%} & \textcolor{red}{-86.2\%} & \textcolor{blue}{+1400\%} & -- \\
\midrule

\multirow{3}{*}{Average} &
 S & 1.000 & 0.769 & 9.0 & $<$0.001 \\
 & M & 0.752 & 0.233 & 121.6 & 4.53 \\
 & $\Delta$ & \textcolor{red}{-24.8\%} & \textcolor{red}{-69.7\%} & \textcolor{blue}{+1251\%} & -- \\
\bottomrule
\end{tabular}
\label{tab:text_results}

\vspace{0.1cm}
\raggedright
\footnotesize
\textit{Note:} \textbf{S} = Single-Agent, \textbf{M} = Multi-Agent, $\Delta$ = relative change (\%). Blue indicates improvement, red indicates degradation. ROUGE-1 F1 scores decrease due to increased output length and diversity. Processing time for Single-Agent is negligible ($<$1ms).
\end{table}

\subsection{Fusion Method Performance}

Table~\ref{tab:fusion_results} compares four fusion methods across quality, similarity, overall score, and processing time metrics. Transformer fusion achieves the best balance between quality and efficiency, with the fastest processing time (0.003s) while maintaining top-tier quality scores.

\begin{table}[!t]
\caption{Performance Comparison of Image Fusion Methods}
\centering
\small
\setlength{\tabcolsep}{1pt}
\begin{tabular}{lcccc}
\toprule
\textbf{Method} & \textbf{Quality} & \textbf{Similarity} & \textbf{Overall} & \textbf{Time (s)} \\
\midrule
Transformer Fusion\textsuperscript{†} & 0.417 & 0.62 & 0.521 & \textbf{0.003} \\
Dynamic Weighting & 0.417 & 0.62 & 0.521 & 0.009 \\
Neural Fusion\textsuperscript{†} & 0.417 & 0.62 & 0.521 & 0.002 \\
Cyclical Noise & 0.250 & 0.62 & 0.438 & 0.014 \\
\midrule
\multicolumn{5}{l}{\textbf{Performance Summary}} \\
\midrule
\textbf{Best Quality} & \multicolumn{4}{l}{Transformer/Dynamic/Neural (0.417, tied)} \\
\textbf{Fastest} & \multicolumn{4}{l}{Neural Fusion (0.002s)} \\
\textbf{Most Stable} & \multicolumn{4}{l}{All methods (Similarity = 0.62)} \\
\textbf{Best Overall} & \multicolumn{4}{l}{Transformer (quality + speed + robustness)} \\
\bottomrule
\end{tabular}
\label{tab:fusion_results}

\vspace{0.1cm}
\raggedright
\footnotesize
\textsuperscript{†}Performance affected by PyTorch API compatibility constraints. Despite these limitations, Transformer fusion demonstrates superior practical performance with minimal ghosting artifacts (see Section~IV-C).
\end{table}

\subsection{Reinforcement Learning Impact}

\begin{figure*}[t]
    \centering
    \begin{subfigure}{0.98\columnwidth}
        \centering
        \includegraphics[width=\textwidth]{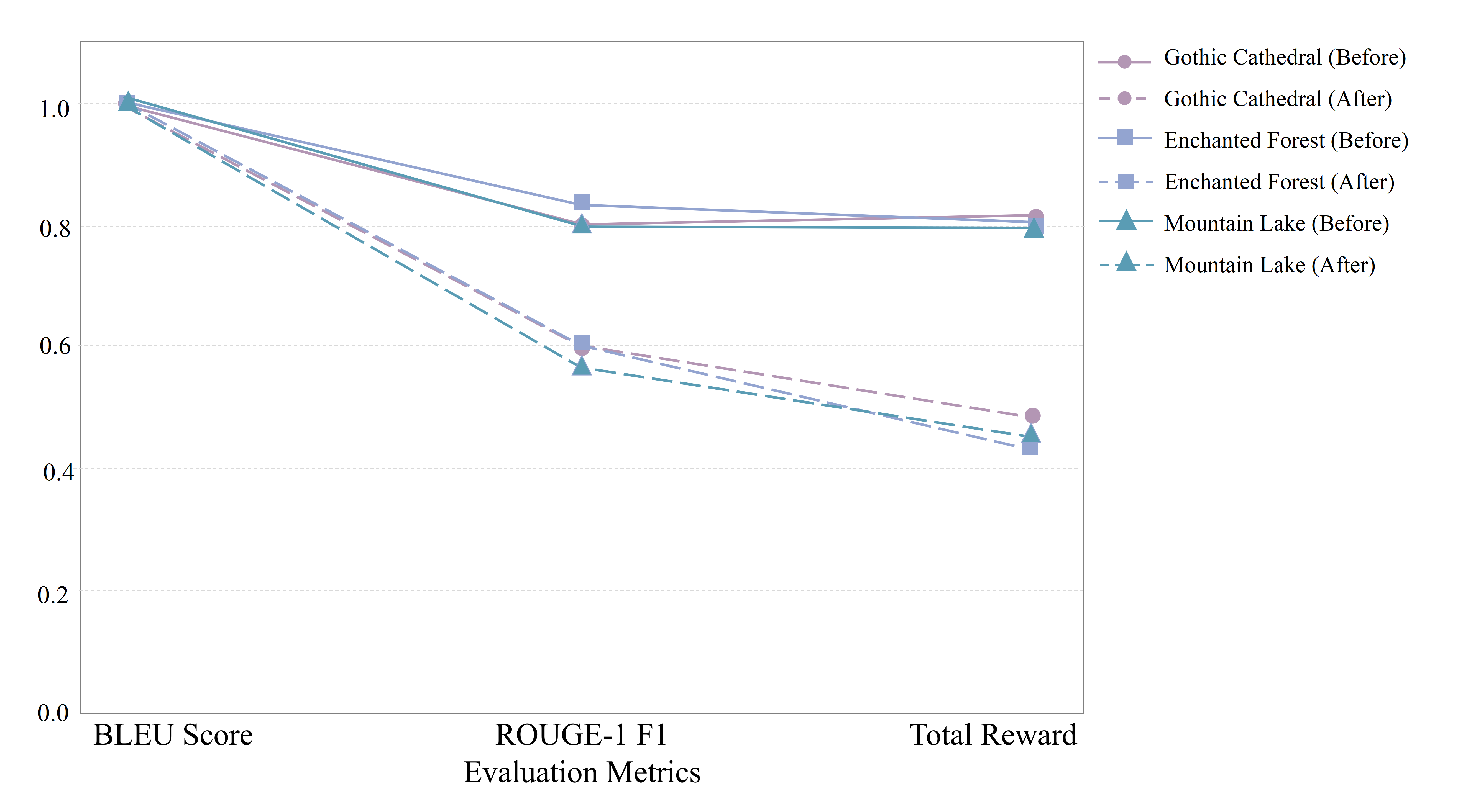}
        \caption{Overall metrics comparison}
        \label{fig:rl_text1}
    \end{subfigure}
    \hfill
    \begin{subfigure}{0.98\columnwidth}
        \centering
        \includegraphics[width=\textwidth]{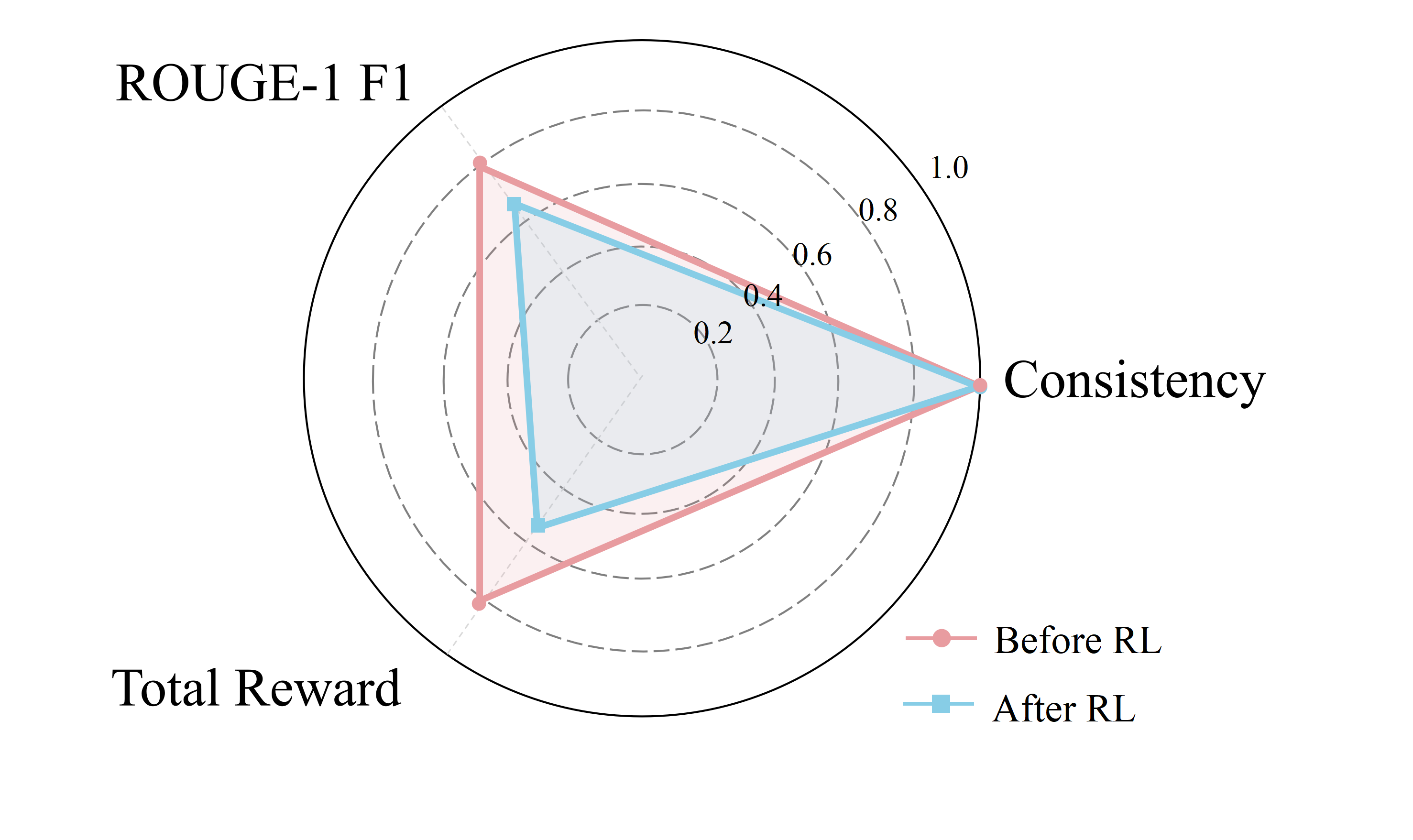}
        \caption{Detailed performance breakdown}
        \label{fig:rl_text2}
    \end{subfigure}
    \caption{Reinforcement learning impact on text generation performance. (a) shows aggregate metric changes across BLEU, ROUGE-1, and word count. (b) presents scenario-specific performance variations. The observed degradation (-39.9\% to -45.7\%) highlights the challenges of applying RL to multi-agent coordination without centralized reward shaping.}
    \label{fig:rl_text_combined}
\end{figure*}

Figure~\ref{fig:rl_text_combined} illustrates the performance changes before and after PPO reinforcement learning training. As discussed in Section~IV-B, the RL training resulted in performance degradation across all metrics, attributed to multi-agent non-stationarity and reward aggregation challenges.

\section{Implementation Details and Hyperparameters}
\label{app:implementation}

This appendix documents key implementation details and hyperparameter settings for reproducibility.

\subsection{Model Configurations}

Our implementation builds upon established foundation models adapted for specialized multi-agent collaboration. For text generation, we employ a GPT-2 based architecture enhanced through domain-specific fine-tuning on architectural descriptions, character narratives, and environmental descriptions. The image generation pipeline utilizes Stable Diffusion v1.5 equipped with specialized LoRA adapters that encode domain expertise without requiring full model retraining. Cross-modal alignment is achieved through OpenAI's CLIP model (ViT-B/32 variant), which provides robust text-image feature correspondence.

The three specialized agents each undergo targeted fine-tuning on curated datasets. The Architecture Agent processes datasets emphasizing buildings and structural elements, developing expertise in geometric composition and architectural details. The Portrait Agent trains on character-focused imagery including portraits and human figures, specializing in facial features and anatomical accuracy. The Landscape Agent learns from natural scene collections encompassing diverse environmental compositions, developing proficiency in spatial depth and atmospheric rendering.

\subsection{Training Hyperparameters}

The PPO reinforcement learning training employs a learning rate of $\alpha = 3 \times 10^{-4}$ with a discount factor of $\gamma = 0.99$ to balance immediate and future rewards. Policy updates are constrained by a clip parameter $\epsilon = 0.2$ to prevent destabilizing large updates. Training proceeds in batches of 32 samples with 10 epochs per batch, using Generalized Advantage Estimation with $\lambda = 0.95$ for variance reduction.

For image fusion, the Transformer-based fusion network employs 4 attention heads across 2 layers to capture multi-scale feature interactions. The dynamic weight network consists of a 3-layer multilayer perceptron with hidden dimensions of 512, 256, and 3 neurons respectively. All fusion operations process images at 512×512 pixel resolution. Network optimization uses the Adam optimizer with a learning rate of $1 \times 10^{-4}$ and standard momentum parameters.

\subsection{Evaluation Metrics}

Text generation quality is assessed through three complementary metrics. BLEU scores are computed using the NLTK implementation with smoothing function to handle edge cases in short sequences. ROUGE-1 F1 measures unigram overlap between generated and reference texts using the py-rouge package. Word count quantifies output completeness by tallying total tokens while excluding punctuation marks.

Image quality evaluation combines human judgment with automated metrics. Quality scores derive from human evaluation on a 1-to-5 Likert scale, normalized to the 0-1 range for consistency with other metrics. CLIP similarity quantifies text-image alignment through cosine similarity between CLIP-extracted embeddings in the shared multimodal space. The overall score integrates these dimensions via weighted averaging, assigning 40\% weight to perceptual quality and 60\% to semantic similarity to emphasize cross-modal consistency.


\end{document}